\documentclass[preprint,a4paper,10pt,oneside,onecolumn]{elsarticle}

\usepackage[T1]{fontenc} % afegit 10-2-2017 (Ramon)

\newif\ifarxiv
\arxivtrue % \arxivtrue to create the arxiv version

\usepackage{lineno,hyperref}
\usepackage{amsmath,amssymb,bm}
\usepackage{graphicx}
\usepackage{multirow}
\usepackage[caption=false]{subfig}
\usepackage{color}
\usepackage{tabularx}

\usepackage{array}
\newcolumntype{C}[1]{>{\centering\let\newline\\\arraybackslash\hspace{0pt}}m{#1}}

\usepackage{float}

\usepackage{siunitx}
\begin{document}

\title{Polysemy and Brevity versus Frequency in Language}

\author[cql]{Bernardino Casas}
\ead{bcasas@cs.upc.edu}
\author[ice]{Antoni Hern\'andez-Fern\'andez}
\ead{antonio.hernandez@upc.edu}
\author[cql]{Neus Catal\`a}
\ead{ncatala@cs.upc.edu}
\author[cql]{Ramon Ferrer-i-Cancho}
\ead{rferrericancho@cs.upc.edu}
\author[cql]{Jaume Baixeries\corref{cor1}}
\ead{jbaixer@cs.upc.edu}

\cortext[cor1]{Corresponding author}

\address[cql]{%
	Complexity \& Quantitative Linguistics Lab,
	Laboratory for Relational Algorithmics,
	Complexity and Learning (LARCA),
	Departament de Ci\`encies de la Computaci\'o,
	Universitat Polit\`ecnica de Catalunya, Barcelona, Catalonia.\\
}

\address[ice]{
	Complexity \& Quantitative Linguistics Lab,
	Laboratory for Relational Algorithmics,
	Complexity and Learning (LARCA),
	Institut de Ci\`encies de l$'$Educaci\'o,
	Universitat Polit\`ecnica de Catalunya,
	Barcelona, Catalonia.\\
}

%%%%%%%%%%%%%%%%%%%%%%%%%%%%%%%%%%%%%%%%%%%%%%%%%%%%%%%%%%%%%%%%%%%%%%%%%%%

\begin{abstract}

% Polysemy, or the capacity for a word to have multiple meanings, has been connected since Zipf's seminal work (1942) with the word-frequency use: more frequent words tend to be the most polysemous. Zipf also defended the law of brevity,  or Zipf's law of Abbreviation (most common words tend to be shorter), and their well-known law for word frequencies that adjust a power-law in the word frequency spectrum.

%In this work we have tested those Zipfian scenarios in the SemCor and the Childes corpora: we have supported Zipfs' hypotheses and we have proved connections between the frequency, the meaning and the brevity of word lemmas.

The pioneering research of G. K. Zipf on the relationship between word frequency and
other word features led to the formulation of various linguistic laws.
The most popular is Zipf's law for word frequencies.
Here we focus on two laws that have been studied less intensively:
the meaning-frequency law, i.e.
the tendency of more frequent words to be more polysemous, and
the law of abbreviation, i.e. the tendency of more frequent words to be shorter.
In a previous work, we tested the robustness of these Zipfian laws for English,
roughly measuring word length in number of characters and
distinguishing adult from child speech.
In the present article, we extend our study to other languages
(Dutch and Spanish) and introduce two additional measures of length: syllabic length and phonemic length.
Our correlation analysis indicates that
both the meaning-frequency law and the law of abbreviation
hold overall in all the analyzed languages.
%\textcolor{blue}{
%FORA: the meaning-frequency law is weaker than the law of abbreviation and
%that length in phonemes strengthens the law of abbreviation specially in English.}

\end{abstract}

\begin{keyword}
Zipf's laws \sep polysemy \sep brevity \sep word frequency
\end{keyword}

%%%%%%%%%%%%%%%%%%%%%%%%%%%%%%%%%%%%%%%%%%%%%%%%%%%%%%%%%%%%%%%%%%%%%%%%%%%

\maketitle

%introduction
%theory/calculation
%results
%discussion
%conclusions
%vitae

\newcommand{\corrEngCHILDES}{
\begin{table}
	\centering
\resizebox{\columnwidth}{!}{ 
	\begin{tabular}{|c||c|c||c|c||c|c||r|}
		\hline
		\multirow{2}{*}{Role} & \multicolumn{2}{c||}{Pearson} & \multicolumn{2}{c||}{Spearman} & \multicolumn{2}{c||}{Kendall} & \multirow{2}{*}{size} \\
		\cline{2-7}
		 & $r$ & p-value & $\rho$ & p-value & $\tau$ & p-value & \\
		 \hline
\hline
& \multicolumn{7}{c|}{\textbf{ CHILDES frequency vs Wordnet polysemy }} \\
\hline
Children &  \textbf{0.059}  &  $ < 10^{-8}$  &  \textbf{0.249}  &  $ < 10^{-140}$  &  \textbf{0.182}  &  $ < 10^{-323}$ &  9,930\\
\hline
Adults &  \textbf{0.09}  &  $ < 10^{-323}$  &  \textbf{0.264}  &  $ < 10^{-257}$  &  \textbf{0.196}  &  $ < 10^{-323}$ &  16,235\\
\hline
\hline
& \multicolumn{7}{c|}{\textbf{ CHILDES frequency vs Number of characters }} \\
\hline
Children &  \textbf{-0.122}  &  $ < 10^{-33}$  &  \textbf{-0.27}  &  $ < 10^{-164}$  &  \textbf{-0.202}  &  $ < 10^{-164}$ &  9,930\\
\hline
Adults &  \textbf{-0.115}  &  $ < 10^{-48}$  &  \textbf{-0.324}  &  $ < 10^{-323}$  &  \textbf{-0.243}  &  $ < 10^{-323}$ &  16,235\\
\hline
\hline
& \multicolumn{7}{c|}{\textbf{ CHILDES frequency vs Number of phonemes }} \\
\hline
Children &  \textbf{-0.123}  &  $ < 10^{-29}$  &  \textbf{-0.31}  &  $ < 10^{-188}$  &  \textbf{-0.234}  &  $ < 10^{-185}$ &  8,547\\
\hline
Adults &  \textbf{-0.11}  &  $ < 10^{-38}$  &  \textbf{-0.361}  &  $ < 10^{-323}$  &  \textbf{-0.273}  &  $ < 10^{-323}$ &  14,146\\
\hline
\hline
& \multicolumn{7}{c|}{\textbf{ CHILDES frequency vs Number of syllables }} \\
\hline
Children &  \textbf{-0.077}  &  $ < 10^{-11}$  &  \textbf{-0.239}  &  $ < 10^{-110}$  &  \textbf{-0.193}  &  $ < 10^{-108}$ &  8,547\\
\hline
Adults &  \textbf{-0.075}  &  $ < 10^{-18}$  &  \textbf{-0.303}  &  $ < 10^{-298}$  &  \textbf{-0.243}  &  $ < 10^{-290}$ &  14,146\\
\hline
	\end{tabular}
	}
\caption{Analysis of the correlations in English with CHILDES frequency. For every correlation test the value of the statistic ($r$, $\rho$ and $\tau$) an the corresponding p-value is shown. Significant correlations are indicated in bold. }
\label{corENGchi}
\end{table}
}

\newcommand{\corrEngWiki}{
\begin{table}
	\centering
\resizebox{\columnwidth}{!}{ 
	\begin{tabular}{|c||c|c||c|c||c|c||r|}
		\hline
		\multirow{2}{*}{Role} & \multicolumn{2}{c||}{Pearson} & \multicolumn{2}{c||}{Spearman} & \multicolumn{2}{c||}{Kendall} & \multirow{2}{*}{size} \\
		\cline{2-7}
		 & $r$ & p-value & $\rho$ & p-value & $\tau$ & p-value & \\
		\hline
\hline
& \multicolumn{7}{c|}{\textbf{ Wikipedia frequency vs Wordnet polysemy }} \\
\hline
Children &  \textbf{0.059}  &  $ < 10^{-8}$  &  \textbf{0.415}  &  $ < 10^{-323}$  &  \textbf{0.301}  &  $ < 10^{-323}$ &  9,975\\
\hline
Adults &  \textbf{0.068}  &  $ < 10^{-323}$  &  \textbf{0.422}  &  $ < 10^{-323}$  &  \textbf{0.307}  &  $ < 10^{-323}$ &  16,286\\
\hline
\hline
& \multicolumn{7}{c|}{\textbf{ Wikipedia frequency vs Number of characters }} \\
\hline
Children &  \textbf{-0.106}  &  $ < 10^{-25}$  &  \textbf{-0.241}  &  $ < 10^{-131}$  &  \textbf{-0.172}  &  $ < 10^{-127}$ &  9,975\\
\hline
Adults &  \textbf{-0.094}  &  $ < 10^{-32}$  &  \textbf{-0.2}  &  $ < 10^{-146}$  &  \textbf{-0.142}  &  $ < 10^{-143}$ &  16,286\\
\hline
\hline
& \multicolumn{7}{c|}{\textbf{ Wikipedia frequency vs Number of phonemes }} \\
\hline
Children &  \textbf{-0.1}  &  $ < 10^{-19}$  &  \textbf{-0.242}  &  $ < 10^{-113}$  &  \textbf{-0.176}  &  $ < 10^{-112}$ &  8,548\\
\hline
Adults &  \textbf{-0.084}  &  $ < 10^{-23}$  &  \textbf{-0.171}  &  $ < 10^{-92}$  &  \textbf{-0.122}  &  $ < 10^{-91}$ &  14,149\\
\hline
\hline
& \multicolumn{7}{c|}{\textbf{ Wikipedia frequency vs Number of syllables }} \\
\hline
Children &  \textbf{-0.06}  &  $ < 10^{-7}$  &  \textbf{-0.17}  &  $ < 10^{-55}$  &  \textbf{-0.131}  &  $ < 10^{-54}$ &  8,548\\
\hline
Adults &  \textbf{-0.054}  &  $ < 10^{-10}$  &  \textbf{-0.102}  &  $ < 10^{-33}$  &  \textbf{-0.078}  &  $ < 10^{-32}$ &  14,149\\
\hline
	\end{tabular}
	}
\caption{Analysis of the correlations in English with Wikipedia frequency. The format is the same as in Table \ref{corENGchi}. }
\label{corENGwik}
\end{table}
}

\newcommand{\corrNldCHILDES}{
	\begin{table}
		\centering
\resizebox{\columnwidth}{!}{ 
		\begin{tabular}{|c||c|c||c|c||c|c||r|}
			\hline
		\multirow{2}{*}{Role} & \multicolumn{2}{c||}{Pearson} & \multicolumn{2}{c||}{Spearman} & \multicolumn{2}{c||}{Kendall} & \multirow{2}{*}{size} \\
		\cline{2-7}
		 & $r$ & p-value & $\rho$ & p-value & $\tau$ & p-value & \\
			\hline
\hline
& \multicolumn{7}{|c|}{\textbf{ CHILDES frequency vs Wordnet polysemy }} \\
\hline
Children &  0.017  &  0.376  &  \textbf{0.19}  &  $ < 10^{-22}$  &  \textbf{0.147}  &  $ < 10^{-323}$ &  2,627\\
\hline
Adults &  0.013  &  0.342  &  \textbf{0.19}  &  $ < 10^{-43}$  &  \textbf{0.149}  &  $ < 10^{-323}$ &  5,273\\
\hline
\hline
& \multicolumn{7}{|c|}{\textbf{ CHILDES frequency vs Number of characters }} \\
\hline
Children &  \textbf{-0.13}  &  $ < 10^{-10}$  &  \textbf{-0.187}  &  $ < 10^{-21}$  &  \textbf{-0.138}  &  $ < 10^{-21}$ &  2,627\\
\hline
Adults &  \textbf{-0.113}  &  $ < 10^{-15}$  &  \textbf{-0.347}  &  $ < 10^{-148}$  &  \textbf{-0.259}  &  $ < 10^{-143}$ &  5,273\\
\hline
\hline
& \multicolumn{7}{|c|}{\textbf{ CHILDES frequency vs Number of phonemes }} \\
\hline
Children &  \textbf{-0.136}  &  $ < 10^{-11}$  &  \textbf{-0.2}  &  $ < 10^{-23}$  &  \textbf{-0.149}  &  $ < 10^{-23}$ &  2,575\\
\hline
Adults &  \textbf{-0.114}  &  $ < 10^{-15}$  &  \textbf{-0.358}  &  $ < 10^{-155}$  &  \textbf{-0.271}  &  $ < 10^{-151}$ &  5,179\\
\hline
\hline
& \multicolumn{7}{|c|}{\textbf{ CHILDES frequency vs Number of syllables }} \\
\hline
Children &  \textbf{-0.1}  &  $ < 10^{-6}$  &  \textbf{-0.169}  &  $ < 10^{-17}$  &  \textbf{-0.134}  &  $ < 10^{-17}$ &  2,575\\
\hline
Adults &  \textbf{-0.093}  &  $ < 10^{-10}$  &  \textbf{-0.323}  &  $ < 10^{-125}$  &  \textbf{-0.258}  &  $ < 10^{-120}$ &  5,179\\
\hline
	\end{tabular}
	}
\caption{Analysis of the correlations in Dutch with CHILDES frequency. The format is the same as in Table \ref{corENGchi}. }
\label{corDUTchi}
\end{table}
}

\newcommand{\corrNldWiki}{
	\begin{table}
		\centering
\resizebox{\columnwidth}{!}{ 
		\begin{tabular}{|c||c|c||c|c||c|c||r|}
			\hline
		\multirow{2}{*}{Role} & \multicolumn{2}{c||}{Pearson} & \multicolumn{2}{c||}{Spearman} & \multicolumn{2}{c||}{Kendall} & \multirow{2}{*}{size} \\
		\cline{2-7}
		 & $r$ & p-value & $\rho$ & p-value & $\tau$ & p-value & \\
			\hline
\hline
& \multicolumn{7}{|c|}{\textbf{ Wikipedia frequency vs Wordnet polysemy }} \\
\hline
Children &  0.002  &  0.931  &  \textbf{0.377}  &  $ < 10^{-89}$  &  \textbf{0.286}  &  $ < 10^{-323}$ &  2,634\\
\hline
Adults &  0.008  &  0.574  &  \textbf{0.395}  &  $ < 10^{-196}$  &  \textbf{0.302}  &  $ < 10^{-323}$ &  5,289\\
\hline
\hline
& \multicolumn{7}{|c|}{\textbf{ Wikipedia frequency vs Number of characters }} \\
\hline
Children &  \textbf{-0.079}  &  $ < 10^{-4}$  &  \textbf{-0.4}  &  $ < 10^{-100}$  &  \textbf{-0.285}  &  $ < 10^{-94}$ &  2,634\\
\hline
Adults &  \textbf{-0.07}  &  $ < 10^{-6}$  &  \textbf{-0.347}  &  $ < 10^{-148}$  &  \textbf{-0.244}  &  $ < 10^{-140}$ &  5,289\\
\hline
\hline
& \multicolumn{7}{|c|}{\textbf{ Wikipedia frequency vs Number of phonemes }} \\
\hline
Children &  \textbf{-0.077}  &  $ < 10^{-4}$  &  \textbf{-0.389}  &  $ < 10^{-93}$  &  \textbf{-0.282}  &  $ < 10^{-87}$ &  2,579\\
\hline
Adults &  \textbf{-0.067}  &  $ < 10^{-5}$  &  \textbf{-0.315}  &  $ < 10^{-119}$  &  \textbf{-0.224}  &  $ < 10^{-114}$ &  5,190\\
\hline
\hline
& \multicolumn{7}{|c|}{\textbf{ Wikipedia frequency vs Number of syllables }} \\
\hline
Children &  \textbf{-0.06}  &  0.002  &  \textbf{-0.363}  &  $ < 10^{-80}$  &  \textbf{-0.28}  &  $ < 10^{-75}$ &  2,579\\
\hline
Adults &  \textbf{-0.055}  &  $ < 10^{-4}$  &  \textbf{-0.281}  &  $ < 10^{-94}$  &  \textbf{-0.212}  &  $ < 10^{-90}$ &  5,190\\
\hline
	\end{tabular}
	}
\caption{Analysis of the correlations in Dutch with Wikipedia frequency. The format is the same as in Table \ref{corENGchi}. }
\label{corDUTwik}
\end{table}
}

\newcommand{\corrSpaCHILDES}{
	\begin{table}
		\centering
\resizebox{\columnwidth}{!}{ 
		\begin{tabular}{|c||c|c||c|c||c|c||r|}
			\hline
		\multirow{2}{*}{Role} & \multicolumn{2}{c||}{Pearson} & \multicolumn{2}{c||}{Spearman} & \multicolumn{2}{c||}{Kendall} & \multirow{2}{*}{size} \\
		\cline{2-7}
		 & $r$ & p-value & $\rho$ & p-value & $\tau$ & p-value & \\
			\hline
\hline
& \multicolumn{7}{|c|}{\textbf{ CHILDES frequency vs Wordnet polysemy }} \\
\hline
Children &  -0.01  &  0.564  &  \textbf{0.162}  &  $ < 10^{-21}$  &  \textbf{0.12}  &  $ < 10^{-323}$ &  3,520\\
\hline
Adults &  0.011  &  0.523  &  \textbf{0.152}  &  $ < 10^{-17}$  &  \textbf{0.113}  &  $ < 10^{-323}$ &  3,217\\
\hline
\hline
& \multicolumn{7}{|c|}{\textbf{ CHILDES frequency vs Number of characters }} \\
\hline
Children &  \textbf{-0.125}  &  $ < 10^{-13}$  &  \textbf{-0.367}  &  $ < 10^{-111}$  &  \textbf{-0.276}  &  $ < 10^{-108}$ &  3,520\\
\hline
Adults &  \textbf{-0.136}  &  $ < 10^{-13}$  &  \textbf{-0.362}  &  $ < 10^{-99}$  &  \textbf{-0.272}  &  $ < 10^{-96}$ &  3,217\\
\hline
\hline
& \multicolumn{7}{|c|}{\textbf{ CHILDES frequency vs Number of phonemes }} \\
\hline
Children &  \textbf{-0.114}  &  $ < 10^{-10}$  &  \textbf{-0.361}  &  $ < 10^{-107}$  &  \textbf{-0.271}  &  $ < 10^{-104}$ &  3,512\\
\hline
Adults &  \textbf{-0.136}  &  $ < 10^{-14}$  &  \textbf{-0.362}  &  $ < 10^{-99}$  &  \textbf{-0.272}  &  $ < 10^{-96}$ &  3,207\\
\hline
\hline
& \multicolumn{7}{|c|}{\textbf{ CHILDES frequency vs Number of syllables }} \\
\hline
Children &  \textbf{-0.109}  &  $ < 10^{-10}$  &  \textbf{-0.344}  &  $ < 10^{-97}$  &  \textbf{-0.276}  &  $ < 10^{-94}$ &  3,520\\
\hline
Adults &  \textbf{-0.126}  &  $ < 10^{-12}$  &  \textbf{-0.354}  &  $ < 10^{-94}$  &  \textbf{-0.284}  &  $ < 10^{-91}$ &  3,217\\
\hline
		\end{tabular}
		}
		\caption{Analysis of the correlations in Spanish with CHILDES frequency. The format is the same as in Table \ref{corENGchi}. }
		\label{corSPAchi}
	\end{table}
}

\newcommand{\corrSpaWiki}{
	\begin{table}
		\centering
\resizebox{\columnwidth}{!}{ 
		\begin{tabular}{|c||c|c||c|c||c|c||r|}
			\hline
		\multirow{2}{*}{Role} & \multicolumn{2}{c||}{Pearson} & \multicolumn{2}{c||}{Spearman} & \multicolumn{2}{c||}{Kendall} & \multirow{2}{*}{size} \\
		\cline{2-7}
		 & $r$ & p-value & $\rho$ & p-value & $\tau$ & p-value & \\
			\hline
\hline
& \multicolumn{7}{|c|}{\textbf{ Wikipedia frequency vs Wordnet polysemy }} \\
\hline
Children &  -0.003  &  0.849  &  \textbf{0.385}  &  $ < 10^{-124}$  &  \textbf{0.282}  &  $ < 10^{-120}$ &  3,524\\
\hline
Adults &  -0.007  &  0.698  &  \textbf{0.359}  &  $ < 10^{-97}$  &  \textbf{0.262}  &  $ < 10^{-95}$ &  3,220\\
\hline
\hline
& \multicolumn{7}{|c|}{\textbf{ Wikipedia frequency vs Number of characters }} \\
\hline
Children &  \textbf{-0.085}  &  $ < 10^{-6}$  &  \textbf{-0.144}  &  $ < 10^{-16}$  &  \textbf{-0.103}  &  $ < 10^{-16}$ &  3,524\\
\hline
Adults &  \textbf{-0.088}  &  $ < 10^{-6}$  &  \textbf{-0.167}  &  $ < 10^{-20}$  &  \textbf{-0.119}  &  $ < 10^{-20}$ &  3,220\\
\hline
\hline
& \multicolumn{7}{|c|}{\textbf{ Wikipedia frequency vs Number of phonemes }} \\
\hline
Children &  \textbf{-0.078}  &  $ < 10^{-5}$  &  \textbf{-0.122}  &  $ < 10^{-12}$  &  \textbf{-0.087}  &  $ < 10^{-12}$ &  3,516\\
\hline
Adults &  \textbf{-0.081}  &  $ < 10^{-5}$  &  \textbf{-0.145}  &  $ < 10^{-15}$  &  \textbf{-0.103}  &  $ < 10^{-15}$ &  3,210\\
\hline
\hline
& \multicolumn{7}{|c|}{\textbf{ Wikipedia frequency vs Number of syllables }} \\
\hline
Children &  \textbf{-0.079}  &  $ < 10^{-5}$  &  \textbf{-0.139}  &  $ < 10^{-15}$  &  \textbf{-0.107}  &  $ < 10^{-16}$ &  3,524\\
\hline
Adults &  \textbf{-0.082}  &  $ < 10^{-5}$  &  \textbf{-0.165}  &  $ < 10^{-20}$  &  \textbf{-0.126}  &  $ < 10^{-20}$ &  3,220\\
\hline
		\end{tabular}
		}
		\caption{Analysis of the correlations in Spanish with Wikipedia frequency. The format is the same as in Table \ref{corENGchi}. }
		\label{corSPAwik}
	\end{table}
}

\newcolumntype{P}[1]{>{\centering\arraybackslash}m{#1}}

\newcommand{\statsWN}{
\begin{table}[htp]
\centering
	\begin{tabular}{|c|P{2.6cm}|P{2.6cm}|P{2.6cm}|}
\hline
\emph{WordNet} & \textbf{Princeton WordNet} \cite{FellBaum1998a} & \textbf{Open Dutch Wordnet} \cite{Postma:Miltenburg:Segers:Schoen:Vossen:2016} & \textbf{Multilingual Central Repository} \cite{Gonzalez-Agirre:Laparra:Rigau:2012} \\
\hline
\emph{Language} & English & Dutch & Spanish \\
\hline
\emph{Synsets} & 117,659 & 30,177 & 38,512 \\
\hline
\emph{Words} & 147,306 & 43,077 & 36,681 \\
\hline
\emph{Core} & 100\% & 67\% & 71\% \\
\hline
\emph{Max} & 75 & 15 & 34 \\
\hline
%\emph{Max 2} & 72 & 15 & 34 \\
%\hline
\emph{Mean} & 1.546 & 1.404 & 1.585\\
\hline
\emph{Median} & 1 & 1 & 1 \\
\hline
\emph{STD} & 1.913 & 0.972 & 1.797 \\
\hline
%\emph{Zero} & 0 & 3,295 & 8,085 \\
%\hline
	\end{tabular}
	\caption{Statistics on used wordnets in Open Multilingual Wordnet to calculate the WordNet polysemy. \emph{Synsets}: Total number of synsets. \emph{Words}: Total number of words. \emph{Core}: the percentage of synsets covered from "core" word senses in Princeton WordNet (approximately the 5000 most frequently used word senses). \emph{Max}: Maximum number of synsets per word. 
	%\emph{Max 2}: Second maximum number of synsets per word. 
	\emph{Mean}: Mean number of synsets per word. \emph{Median}: Median of the number of synsets per word. \emph{STD}: Standard Deviation. 
	%\emph{Zero}: Words included in WordNet that don't have associated any synset.
	}
	\label{tab:statsWN}
\end{table}
}

\newcommand{\analyzedWords}{
	\begin{table}[ht]
		\centering
		\resizebox{\textwidth}{!}{
		\begin{tabular}{|c||c||r|r|c||r|r|c|}
			\hline
			\multirow{2}{*}{\textbf{Lang.}} & \multirow{2}{*}{\textbf{Role}} & \multicolumn{3}{c||}{\textbf{Types}} & \multicolumn{3}{c|}{\textbf{Tokens}} \\
			\cline{3-8}
			 & & analyzed & total & cover & analyzed & total & cover \\
			\hline
\hline
\multirow{2}{*}{English} & Children & 9,930 & 29,017 & 34\% & 1,596,726 & 2,308,675 & 69\% \\
\cline{2-8}
& Adults & 16,235 & 25,135 & 64\% & 3,008,148 & 4,584,213 & 65\% \\
\hline
\hline
\multirow{2}{*}{Dutch} & Children & 2,627 & 13,666 & 19\% & 313,556 & 628,622 & 50\% \\
\cline{2-8}
& Adults & 5,273 & 19,037 & 28\% & 1,008,393 & 2,122,354 & 48\% \\
\hline
\hline
\multirow{2}{*}{Spanish} & Children & 3,520 & 27,167 & 13\% & 253,145 & 864,603 & 29\% \\
\cline{2-8}
 & Adults & 3,217 & 19,975 & 16\% & 288,660 & 997,901 & 29\% \\
\hline
		\end{tabular}}
		\caption{Number of analyzed and total words (by types and tokens) obtained
			from CHILDES conversations for each language and role. \emph{cover}: percentage of words (by types or tokens) that appear in the corresponding WordNet and Wikipedia lexicon.}
		\label{tab:analyzed-words}
	\end{table}
}

\newcommand{\lletresphon}{ 
	\begin{center} 
		\begin{table}[H] 
			\resizebox{\columnwidth}{!}{ 
				\begin{tabular}{|c||cc|cc||cc|cc||c|} 
					\hline
					\multirow{3}{*}{Role} & \multicolumn{4}{c||}{CHILDES Frequency} & \multicolumn{4}{c||}{Wikipedia Frequency} &  \multirow{3}{*}{size} \\ 
					\cline{2-9} 
					& \multicolumn{2}{c|}{Pearson} & \multicolumn{2}{c||}{Spearman} 
					& \multicolumn{2}{c|}{Pearson} & \multicolumn{2}{c||}{Spearman} & size \\ 
					\cline{2-9} 
					& t & p-value & t & p-value & t & p-value & t & p-value & \\ 
					\hline 
					\multicolumn{10}{|c|}{\textbf{English}} \\ 
					\hline 
					Children  & \textbf{ -2.484 } &  0.013 & 0.022  &  0.982 & \textbf{ -3.693 } &  $ < 10^{-3}$ & \textbf{ -4.572 } &  $ < 10^{-5}$ &  8,548   \\
					\hline
					Adults  & \textbf{ -3.562 } &  $ < 10^{-3}$ & \textbf{ 2.501 } &  0.012 & \textbf{ -4.553 } &  $ < 10^{-5}$ & \textbf{ -8.837 } &  $ < 10^{-17}$ &  14,149   \\
					\hline
					\multicolumn{10}{|c|}{\textbf{Dutch}} \\ 
					\hline 
					Children  & 0.395  &  0.693 & 1.417  &  0.157 & -0.472  &  0.637 & -1.049  &  0.294 &  2,579   \\
					\hline
					Adults  & -0.251  &  0.802 & 1.624  &  0.104 & -0.793  &  0.428 & \textbf{ -6.491 } &  $ < 10^{-10}$ &  5,190   \\
					\hline
					\multicolumn{10}{|c|}{\textbf{Spanish}} \\ 
					\hline 
					Children  & -0.861  &  0.389 & -0.391  &  0.696 & -1.203  &  0.229 & \textbf{ -3.596 } &  $ < 10^{-3}$ &  3,516   \\
					\hline
					Adults  & 0.098  &  0.922 & 0.571  &  0.568 & -1.194  &  0.233 & \textbf{ -3.43 } &  $ < 10^{-3}$ &  3,210   \\
					\hline
				\end{tabular} 
			} 
			\caption{Steiger's test between variables \textit{Number of characters} and \textit{Number of phonemes}.
				The test indicates if the difference between the $r_1$ and $r_2$ is significant. 
				$r_1$ is the correlation between \textit{Frequency} and \textit{Number of characters} 
				and $r_2$ is the correlation between \textit{Frequency} and \textit{Number of phonemes}. 
				\textit{Frequency} is the shared variable. 
				The table has two parts: one for the results when the CHILDES frequency
				is considered the shared variable, and another 
				when Wikipedia frequency is the shared variable. Within each part,  
				the \textbf{t} statistic of Steiger test and \textbf{p-value}, 
				the corresponding p-value, are shown for Pearson and Spearman correlations. 
				Significant results are in bold.} 
			\label{nlletresnphon} 
		\end{table} 
	\end{center} 
}

\newcommand{\lletressyllab}{ 
	\begin{center} 
		\begin{table}[H] 
			\resizebox{\columnwidth}{!}{ 
				\begin{tabular}{|c||cc|cc||cc|cc||c|} 
					\hline
					\multirow{3}{*}{Role} & \multicolumn{4}{c||}{CHILDES Frequency} & \multicolumn{4}{c||}{Wikipedia Frequency} &  \multirow{3}{*}{size} \\ 
					\cline{2-9} 
					& \multicolumn{2}{c|}{Pearson} & \multicolumn{2}{c||}{Spearman} 
					& \multicolumn{2}{c|}{Pearson} & \multicolumn{2}{c||}{Spearman} & size \\ 
					\cline{2-9} 
					& t & p-value & t & p-value & t & p-value & t & p-value & \\ 
					\hline 
					\multicolumn{10}{|c|}{\textbf{English}} \\ 
					\hline 
					Children  & \textbf{ -7.922 } &  $ < 10^{-14}$ & \textbf{ -9.812 } &  $ < 10^{-21}$ & \textbf{ -7.895 } &  $ < 10^{-14}$ & \textbf{ -13.23 } &  $ < 10^{-38}$ &  8,548   \\
					\hline
					Adults  & \textbf{ -9.045 } &  $ < 10^{-18}$ & \textbf{ -9.3 } &  $ < 10^{-19}$ & \textbf{ -8.817 } &  $ < 10^{-17}$ & \textbf{ -19.092 } &  $ < 10^{-79}$ &  14,149   \\
					\hline
					\multicolumn{10}{|c|}{\textbf{Dutch}} \\ 
					\hline 
					Children  & \textbf{ -3.277 } &  0.001 & \textbf{ -2.051 } &  0.040 & \textbf{ -2.01 } &  0.045 & \textbf{ -3.519 } &  $ < 10^{-3}$ &  2,579   \\
					\hline
					Adults  & \textbf{ -3.079 } &  0.002 & \textbf{ -4.199 } &  $ < 10^{-4}$ & \textbf{ -2.164 } &  0.031 & \textbf{ -9.394 } &  $ < 10^{-20}$ &  5,190   \\
					\hline
					\multicolumn{10}{|c|}{\textbf{Spanish}} \\ 
					\hline 
					Children  & -1.775  &  0.076 & \textbf{ -2.704 } &  0.007 & -0.701  &  0.483 & -0.491  &  0.623 &  3,524   \\
					\hline
					Adults  & -1.139  &  0.255 & -1.006  &  0.314 & -0.691  &  0.490 & -0.28  &  0.780 &  3,220   \\
					\hline
				\end{tabular} 
			} 
			\caption{Steiger's test between variables \textit{Number of characters} and \textit{Number of syllables}.
				The format is the same as in Table \ref{nlletresnphon}.} 
			\label{nlletresnsyllab} 
		\end{table} 
	\end{center} 
}

\newcommand{\phonsyllab}{ 
	\begin{center} 
		\begin{table}[H] 
			\resizebox{\columnwidth}{!}{ 
				\begin{tabular}{|c||cc|cc||cc|cc||c|} 
					\hline
					\multirow{3}{*}{Role} & \multicolumn{4}{c||}{CHILDES Frequency} & \multicolumn{4}{c||}{Wikipedia Frequency} &  \multirow{3}{*}{size} \\ 
					\cline{2-9} 
					& \multicolumn{2}{c|}{Pearson} & \multicolumn{2}{c||}{Spearman} 
					& \multicolumn{2}{c|}{Pearson} & \multicolumn{2}{c||}{Spearman} & size \\ 
					\cline{2-9} 
					& t & p-value & t & p-value & t & p-value & t & p-value & \\ 
					\hline 
					\multicolumn{10}{|c|}{\textbf{English}} \\ 
					\hline 
					Children  & \textbf{ -7.145 } &  $ < 10^{-12}$ & \textbf{ -10.844 } &  $ < 10^{-26}$ & \textbf{ -6.135 } &  $ < 10^{-9}$ & \textbf{ -10.937 } &  $ < 10^{-26}$ &  8,548   \\
					\hline
					Adults  & \textbf{ -7.67 } &  $ < 10^{-13}$ & \textbf{ -12.544 } &  $ < 10^{-35}$ & \textbf{ -6.581 } &  $ < 10^{-10}$ & \textbf{ -14.275 } &  $ < 10^{-45}$ &  14,149   \\
					\hline
					\multicolumn{10}{|c|}{\textbf{Dutch}} \\ 
					\hline 
					Children  & \textbf{ -3.488 } &  $ < 10^{-3}$ & \textbf{ -2.901 } &  0.004 & -1.662  &  0.097 & \textbf{ -2.628 } &  0.009 &  2,579   \\
					\hline
					Adults  & \textbf{ -3.007 } &  0.003 & \textbf{ -5.232 } &  $ < 10^{-6}$ & -1.71  &  0.087 & \textbf{ -4.951 } &  $ < 10^{-6}$ &  5,190   \\
					\hline
					\multicolumn{10}{|c|}{\textbf{Spanish}} \\ 
					\hline 
					Children  & -1.093  &  0.274 & \textbf{ -2.413 } &  0.016 & -0.09  &  0.929 & 1.19  &  0.234 &  3,516   \\
					\hline
					Adults  & -1.162  &  0.245 & -1.227  &  0.220 & -0.102  &  0.918 & 1.281  &  0.200 &  3,210   \\
					\hline
				\end{tabular} 
			} 
			\caption{Steiger's test between variables \textit{Number of phonemes} and \textit{Number of syllables}.
				The format is the same as in Table \ref{nlletresnphon}.} 
			\label{nphonnsyllab} 
		\end{table} 
	\end{center} 
}

\newcommand{\taulaSteigerENG}{
	\begin{center}
		\begin{table}[H]
			\resizebox{\columnwidth}{!}{
\begin{tabular}{ |l|l|l|c|c|c| }
\hline
Role & Frequency & Correlation & Char. vs Phon. & Phon. vs Syllables & Char. vs Syllables \\
\hline
\multirow{2}{*}{ Children }  & \multirow{2}{*}{ CHILDES } & Pearson  &  $\boldsymbol{>^*}$  &  $\boldsymbol{>^*}$  &  $\boldsymbol{>^*}$ \\
\cline{3-6}
 & &  Spearman  &  $<$  &  $\boldsymbol{>^*}$  &  $\boldsymbol{>^*}$ \\
\cline{3-6}
 & &  Kendall  &  $<$  &  $>$  &  $>$ \\
\cline{3-6}
\cline{2-6}
 & \multirow{2}{*}{ Wikipedia } & Pearson  &  $\boldsymbol{>^*}$  &  $\boldsymbol{>^*}$  &  $\boldsymbol{>^*}$ \\
\cline{3-6}
 & &  Spearman  &  $\boldsymbol{>^*}$  &  $\boldsymbol{>^*}$  &  $\boldsymbol{>^*}$ \\
\cline{3-6}
 & &  Kendall  &  $>$  &  $>$  &  $>$ \\
\cline{3-6}
\cline{2-6}
\hline
\multirow{2}{*}{ Adults }  & \multirow{2}{*}{ CHILDES } & Pearson  &  $\boldsymbol{>^*}$  &  $\boldsymbol{>^*}$  &  $\boldsymbol{>^*}$ \\
\cline{3-6}
 & &  Spearman  &  $\boldsymbol{<^*}$  &  $\boldsymbol{>^*}$  &  $\boldsymbol{>^*}$ \\
\cline{3-6}
 & &  Kendall  &  $<$  &  $>$  &  $>$ \\
\cline{3-6}
\cline{2-6}
 & \multirow{2}{*}{ Wikipedia } & Pearson  &  $\boldsymbol{>^*}$  &  $\boldsymbol{>^*}$  &  $\boldsymbol{>^*}$ \\
\cline{3-6}
 & &  Spearman  &  $\boldsymbol{>^*}$  &  $\boldsymbol{>^*}$  &  $\boldsymbol{>^*}$ \\
\cline{3-6}
 & &  Kendall  &  $>$  &  $>$  &  $>$ \\
\cline{3-6}
\cline{2-6}
\hline
\end{tabular}
			}
			\caption{For \textbf{English},
			the results of the Steiger's test between a frequency measure
			(CHILDES or Wikipedia)
			and the three different length measures of a word
			(\textit{Char.} for number of characters,
			 \textit{Phon.} for the number of phonemes and
			 \textit{Syllables} for the number of Syllables).
			Column \textit{Role} indicates the subject (children or adults),
			the \textit{Frequency} column indicates the source of the frequency
			measure (the shared variable),
			and the column \textit{Correlation} indicates the type of correlation.
			As for the remaining columns, we have the combination of two length
			measures. The content is $>^*$ when the first length variable is more
			correlated to the frequency measure than the second variable and
			the Steiger's test is significant. If the second measure is more correlated
			than the first and the test is significant, then, the content is $<^*$.
			If the test is not significant, then the contents are $>$ or $<$ depending
			on what variable shows a higher correlation to the frequency measure.}
			\label{ntaulaSteigerENG}
		\end{table}
	\end{center}
}

\newcommand{\taulaSteigerNLD}{
	\begin{center}
		\begin{table}[H]
			\resizebox{\columnwidth}{!}{
\begin{tabular}{ |l|l|l|c|c|c| }
\hline
Role & Frequency & Correlation & Char. vs Phon. & Phon. vs Syllables & Char. vs Syllables \\
\hline
\multirow{2}{*}{ Children }  & \multirow{2}{*}{ CHILDES } & Pearson  &  $<$  &  $\boldsymbol{>^*}$  &  $\boldsymbol{>^*}$ \\
\cline{3-6}
 & &  Spearman  &  $<$  &  $\boldsymbol{>^*}$  &  $\boldsymbol{>^*}$ \\
\cline{3-6}
 & &  Kendall  &  $<$  &  $>$  &  $>$ \\
\cline{3-6}
\cline{2-6}
 & \multirow{2}{*}{ Wikipedia } & Pearson  &  $>$  &  $>$  &  $\boldsymbol{>^*}$ \\
\cline{3-6}
 & &  Spearman  &  $>$  &  $\boldsymbol{>^*}$  &  $\boldsymbol{>^*}$ \\
\cline{3-6}
 & &  Kendall  &  $>$  &  $>$  &  $>$ \\
\cline{3-6}
\cline{2-6}
\hline
\multirow{2}{*}{ Adults }  & \multirow{2}{*}{ CHILDES } & Pearson  &  $>$  &  $\boldsymbol{>^*}$  &  $\boldsymbol{>^*}$ \\
\cline{3-6}
 & &  Spearman  &  $<$  &  $\boldsymbol{>^*}$  &  $\boldsymbol{>^*}$ \\
\cline{3-6}
 & &  Kendall  &  $<$  &  $>$  &  $>$ \\
\cline{3-6}
\cline{2-6}
 & \multirow{2}{*}{ Wikipedia } & Pearson  &  $>$  &  $>$  &  $\boldsymbol{>^*}$ \\
\cline{3-6}
 & &  Spearman  &  $\boldsymbol{>^*}$  &  $\boldsymbol{>^*}$  &  $\boldsymbol{>^*}$ \\
\cline{3-6}
 & &  Kendall  &  $>$  &  $>$  &  $>$ \\
\cline{3-6}
\cline{2-6}
\hline
\end{tabular}
			}
			\caption{Results of the Steiger's test for \textbf{Dutch}.
				The format is the same as in Table \ref{ntaulaSteigerENG}.}
			\label{ntaulaSteigerNLD}
		\end{table}
	\end{center}
}

\newcommand{\taulaSteigerSPA}{
	\begin{center}
		\begin{table}[H]
			\resizebox{\columnwidth}{!}{
\begin{tabular}{ |l|l|l|c|c|c| }
\hline
Role & Frequency & Correlation & Char. vs Phon. & Phon. vs Syllables & Char. vs Syllables \\
\hline
\multirow{2}{*}{ Children }  & \multirow{2}{*}{ CHILDES } & Pearson  &  $>$  &  $>$  &  $>$ \\
\cline{3-6}
 & &  Spearman  &  $>$  &  $\boldsymbol{>^*}$  &  $\boldsymbol{>^*}$ \\
\cline{3-6}
 & &  Kendall  &  $>$  &  $<$  &  $<$ \\
\cline{3-6}
\cline{2-6}
 & \multirow{2}{*}{ Wikipedia } & Pearson  &  $>$  &  $>$  &  $>$ \\
\cline{3-6}
 & &  Spearman  &  $\boldsymbol{>^*}$  &  $<$  &  $>$ \\
\cline{3-6}
 & &  Kendall  &  $>$  &  $<$  &  $<$ \\
\cline{3-6}
\cline{2-6}
\hline
\multirow{2}{*}{ Adults }  & \multirow{2}{*}{ CHILDES } & Pearson  &  $<$  &  $>$  &  $>$ \\
\cline{3-6}
 & &  Spearman  &  $<$  &  $>$  &  $>$ \\
\cline{3-6}
 & &  Kendall  &  $<$  &  $<$  &  $<$ \\
\cline{3-6}
\cline{2-6}
 & \multirow{2}{*}{ Wikipedia } & Pearson  &  $>$  &  $>$  &  $>$ \\
\cline{3-6}
 & &  Spearman  &  $\boldsymbol{>^*}$  &  $<$  &  $>$ \\
\cline{3-6}
 & &  Kendall  &  $>$  &  $<$  &  $<$ \\
\cline{3-6}
\cline{2-6}
\hline
\end{tabular}
			}
			\caption{Results of the Steiger's test for \textbf{Spanish}.
				The format is the same as in Table \ref{ntaulaSteigerENG}.}
			\label{ntaulaSteigerSPA}
		\end{table}
	\end{center}
}

%\graphicspath{{./../../dades/}}
\graphicspath{{figs/}}

\newcommand{\figura}[3]{
\def\ampl{.40} % amplada de les figures
\begin{figure}[htp]
	\centering
	\resizebox{\textwidth}{!}{
	\begin{tabular}{lcc}
		& {\tiny Children} & {\tiny Adults} \\
		{\footnotesize English} & 
		\raisebox{-.5\height}
		{\includegraphics[width=\ampl\textwidth]{#1_#2_eng_childes_CHI_regressio}}
		&
		\raisebox{-.5\height}
		{\includegraphics[width=\ampl\textwidth]{#1_#2_eng_childes_ADU_regressio}}
		\\
		{\footnotesize Dutch} & 
		\raisebox{-.5\height}
		{\includegraphics[width=\ampl\textwidth]{#1_#2_nld_childes_CHI_regressio}}
		&
		\raisebox{-.5\height}
		{\includegraphics[width=\ampl\textwidth]{#1_#2_nld_childes_ADU_regressio}}
		\\
		{\footnotesize Spanish} & 
		\raisebox{-.5\height}
		{\includegraphics[width=\ampl\textwidth]{#1_#2_spa_childes_CHI_regressio}}
		&
		\raisebox{-.5\height}
		{\includegraphics[width=\ampl\textwidth]{#1_#2_spa_childes_ADU_regressio}}
	\end{tabular}}
	\caption{#3}
	\label{fig:#1_vs_#2}
\end{figure}
}

%%%%%%%%%%%%%%%%%%%%%%%%%%%%%%%%%%%%%%%%%%%%%%%%%%%%%%%%%%%%%%%%%%%%%%%%%%%
%

\section{Introduction}
\label{intro}

The linguist George Kingsley Zipf (1902-1950) is known for his investigations
on statistical laws of language \cite{Zipf1949a,Zipf1968a}.
Perhaps the most popular one is
\textbf{Zipf's law for word frequencies} \cite{Zipf1949a,zipf45meaning},
that states that the frequency of the $i$-th most frequent word in a text follows approximately
\begin{equation}
f \propto i^{-\alpha}
\label{Zipfs_law_equation}
\end{equation}
where $f$ is the frequency of that word,
$i$ its rank or order and
$\alpha$ is a constant ($\alpha \approx 1$).
% As in \cite{DBLP:dblp_journals/corr/Ferrer-i-Cancho14e} is pointed out,
Zipf's law is an example of power-law model for the relationship between two variables \cite{Naranan1998}.
Zipf's law for word frequencies can be explained by information theoretic models
of communication \cite{DBLP:dblp_journals/corr/Ferrer-i-Cancho14e}
and is a robust pattern of language that presents invariance
with text length in a sufficiently long text \cite{1367-2630-15-9-093033}, 
and little sensitivity with respect
to the linguistic units considered \cite{10.1371/journal.pone.0129031}.
% He tret la falca publicitaria perque la veig massa forçada aqui. Ja se'n fa publicitat en altres parts de l'article.
% That robustness could be reinforced by the presence of self-organized criticality (SOC) in human voice \cite{luque2015scaling, GonzalezTorre2016}.
The focus of this paper is to test the robustness of two statistical laws in linguistics that have been studied less intensively:
\begin{itemize}
\item \textbf{Meaning-frequency law} \cite{zipf45meaning},
 the tendency of more frequent words to be more polysemous.
% which states that more frequent words tend to be the most polysemous,
% considering that polysemy is the capacity for a word
% to have multiple meanings or more than one related sense.
Zipf predicted that the $m$ number of meanings of a word should follow
\begin{equation}
m \approx f^{\delta}
\end{equation}
where $f$ is the frequency of a word and $\delta \approx 0.5$. Zipf never tested the validity of this equation. He only derived it from Zipf's law for word frequencies (Eq. \ref{Zipfs_law_equation}) and the law of meaning distribution \cite{zipf45meaning} (see \cite{Ferrer2014d} for a general derivation). The latter links $m$ with $i$, namely, the frequency rank. Zipf proposed and tested \cite{Zipf1949a,zipf45meaning}
\begin{equation}
m \propto i^{-\gamma}
\label{law_of_meaning_distribution_equation}
\end{equation}
\noindent
where $\gamma$ is a constant ($\gamma \approx 0.5$).

\item \textbf{Zipf's law of abbreviation} \cite{Zipf1949a,grzybek2006contributions},
the tendency of more frequent words to be shorter or smaller.
In his pioneering research, Zipf made this observation but did not propose any mathematical 
formulae to model that dependency \cite{Zipf1949a}. 
Power-law like functions were suggested 
later on by other researchers (\cite{Strauss2007a}).
\end{itemize}

These laws are examples of laws where the predictor
is the word frequency and the response is another word feature.
These laws are regarded as universal although the only evidence of
their universality is that they hold in every language
where they have been tested so far \cite{Ilgen2007a}.
% no se perque cal citar \cite{Altmann2015a,Ferrer2016compression} aqui (Ramon).
Because of their generality, these laws have triggered
modeling efforts that attempt to explain their origin and support
their presumable universality with the help of abstract mechanisms
or communication principles \cite{Ferrer2012d,Ferrer2017b},
or exploring directly from voice those statistical patterns in levels under the phoneme scale \cite{GonzalezTorre2016}. Therefore, investigating the experimental conditions under which these laws surface is crucial.

In a previous work \cite{pilsen}, we have studied these linguistic laws
in a large corpus of child and adult language (CHILDES) \cite{MacWhinney2000a}.
We extracted semantic polysemy values from WordNet \cite{FellBaum1998a}
and SemCor corpus \footnote{\url{http://multisemcor.fbk.eu/semcor.php}},
and defined word length simply as the number of characters per word.

In this present article we extend our research in \cite{pilsen}
by re-analyzing the behaviour of those linguistic laws in children and
adults separately, using the transcripts in the CHILDES database,
and exploring different definitions of word frequency,
word polysemy and word length.
% \textcolor{red}{However, we will not enter here into the classical distinction of 
% the elements that constitute words according to frequency, 
% which would be the subject of another study \cite{LANDAUER1973119}}.
In order to test the statistical validity
of these linguistic laws, we also expand the number of languages to
Dutch and Spanish, as well as English, which was the only language analyzed in \cite{pilsen}.
%In our analysis, and for all these languages,
%only content words (nouns, verbs, adjectives and adverbs)
%have been taken into account.
Concerning word frequency, we consider two major sources of estimation:
the CHILDES database \cite{MacWhinney2000a}
and Wikipedia \cite{GREFENSTETTE16.624}.
%From these sources we
%Frequency estimates
%from the CHILDES database
%are divided into four types depending on the role of the speaker: child, mother, father and investigator.
Frequency estimates are computed separately for children and adults (comprising mothers, fathers and investigators).
This division allows us to compare children and adults linguistic production:
\textit{motherese} also known as child-directed speech (CDS)
or infant-directed speech (IDS) has been studied for many years
and it is still a hot topic of research \cite{motheresereview2013}.
% Tret perque no podem fer preguntes retòriques de questions que despres no tractem en profunditat tan a nivell d'analisi com de presentacio i discussio de resultats (Ramon).
% So, is there any statistical variation in the use of words by adults when speaking with children?

Concerning polysemy, we define the polysemy of a word as the number of different senses it has,
based on the WordNet of its corresponding language (Princeton WordNet, Open Dutch WordNet
and Multilingual Central Repository for Spanish). Hereafter, we will refer to this polysemy
as WordNet polysemy. We assume that the polysemy measure provided by WordNet does
not distinguish between different types of polysemy and we are aware of
the inherent difficulties of borrowing this conceptual framework
(see \cite{Ide2006,Kilgarriff1992,Zugarramurdi,FRAGA20171}).
Concerning word length, we consider three different units of measurement:
a graphical unit (number of characters) and two phonetic units
(number of phonemes and number of syllables).
%Therefore, the SemCor corpus contains
%SemCor polysemy and SemCor frequency, as well as the length of its lemmas,
%and the CHILDES database contains CHILDES frequency, the length of its lemmas,
%and has been enriched with CELEX frequency, WordNet polysemy, and SemCor polysemy.
%The conditions above lead to $1 + 2 \times 2 = 5$ major ways of investigating the meaning-frequency law and
%to $1 + 2 = 3$ ways of investigating the law of abbreviation (see details in Section \ref{sec:methods}).
%The choice made in this study
%should not be considered a limitation,
%since we plan to extend the range of data sources and measures in future studies
%(we explain these possibilities in Section \ref{discussion}).
From the sources for obtaining word frequency and polysemy values and from the variety of measurement units for word length, we come up to eight major ways of investigating meaning-frequency law and law of abbreviation, for each language under study.
% Tret pel Ramon. Motiu, la seccio discussió es minimalista i no entre en detalls.
% In Section \ref{discussion} we analyze the choice of each word feature and how it affects to each language.

%In this study, we have limited
%the choice of some corpuses as well as the source of some measures.
%However, we are planning to extend the range of data sources and measures.
%We explain these possibilities in Section \ref{discussion}}.

In this paper, we investigate these laws qualitatively using measures of correlation between two variables.
Thus, the law of abbreviation is defined as a significant negative correlation between the frequency of a word and its length for any unit of measurement.
%Besides this, we perform a test for comparing pairs of dependent correlations, i.e.
%pairs of correlations (between each unit to measure word length and frequency)
%that involve a common variable (frequency).
The meaning-frequency law is defined as a significant positive correlation
between the frequency of a word and its WordNet polysemy,
a proxy for the number of meanings of a word.
While our approach to these laws is non-parametric
(we are not assuming any particular model for
the relationship between two variables), traditional
research on statistical laws of language is mostly parametric,
assuming some sort of power law or generalizations of power
laws \cite{Naranan1998,Strauss2007a,Altmann1980a}.

We adopt these correlational definitions to remain agnostic about
the actual functional dependency between the variables,
which is currently under revision for various statistical
laws of language \cite{Altmann2015a,Font2015a,Ferrer2012h}.
We will show that a significant correlation of the right sign
is found in the majority of combinations of conditions mentioned above,
providing support for the hypothesis that
these laws are originated from abstract mechanisms.
We propose as well some hypotheses
to explain why in some exceptional cases
the analyzed variables do not correlate significantly.
% even applying the same methods in all languages.

% ??? The hypothesis about the influence of polysemy on word frequency suggested by Zipf \cite{zipf45meaning} has been proven at the morphological level for derivational affixes and morphemes which are used as verbs, nouns, or adjectives \cite{DBLP:journals/jql/Krott99}.

The remainder of the article is organized as follows.
Section 2 revises the power law model
that Zipf proposed for the law of meaning distribution \cite{Zipf1949a,zipf45meaning}
to illustrate the challenges of the parametric approach presented here.
Then we justify the convenience of a non-parametric approach
(and correlation analysis) that we have adopted in this
article for statistical laws of language involving word frequency.
Sections \ref{materials} and \ref{sec:methods} present, respectively,
the materials (databases) and the methods employed to analyze them.
Section \ref{results} presents the results of our analysis of
the meaning-frequency law and the law of abbreviation.
Section \ref{discussion} discusses our findings and suggests future work.

\section{Revisiting Zipf's law of meaning distribution}

\label{law_of_meaning_distribution_section}

To check if the equation that Zipf proposed for the law of meaning distribution (Eq. \ref{law_of_meaning_distribution_equation}) holds on modern corpora, we have reproduced
the computations exposed in \cite{Zipf1949a}
%\textit{Human Behaviour and the Principle of Least Effort} by G. Zipf,
on a data set that we explore in depth in the next sections.

When plotting the relationship between number of meanings per word (on the ordinate)
and frequency rank (abscissa), Zipf applied a linear binning technique to reduce noise. When using bins of length $\lambda$, the 1st bin is formed by the $\lambda$ most frequent words, the 2nd bin is formed by the next $\lambda$ most frequent words,... etc.
Formally, the $j$-th bin is defined by words whose rank $i$ satisfies
\begin{equation}
\lambda (j-1) + 1 \leq i \leq \lambda j.
\end{equation}
Zipf plotted the relationship between the average
number of meanings of the $j$-th bin and $j$
\cite[p. 30]{Zipf1949a} and fitted a power law
(Eq. \ref{law_of_meaning_distribution_equation}).
We follow the same method for a sample from
the CHILDES database (see Section
\ref{sub:childes} for further details).

\begin{figure}[htp]
\centering
\ifarxiv
	\includegraphics[width=\textwidth]{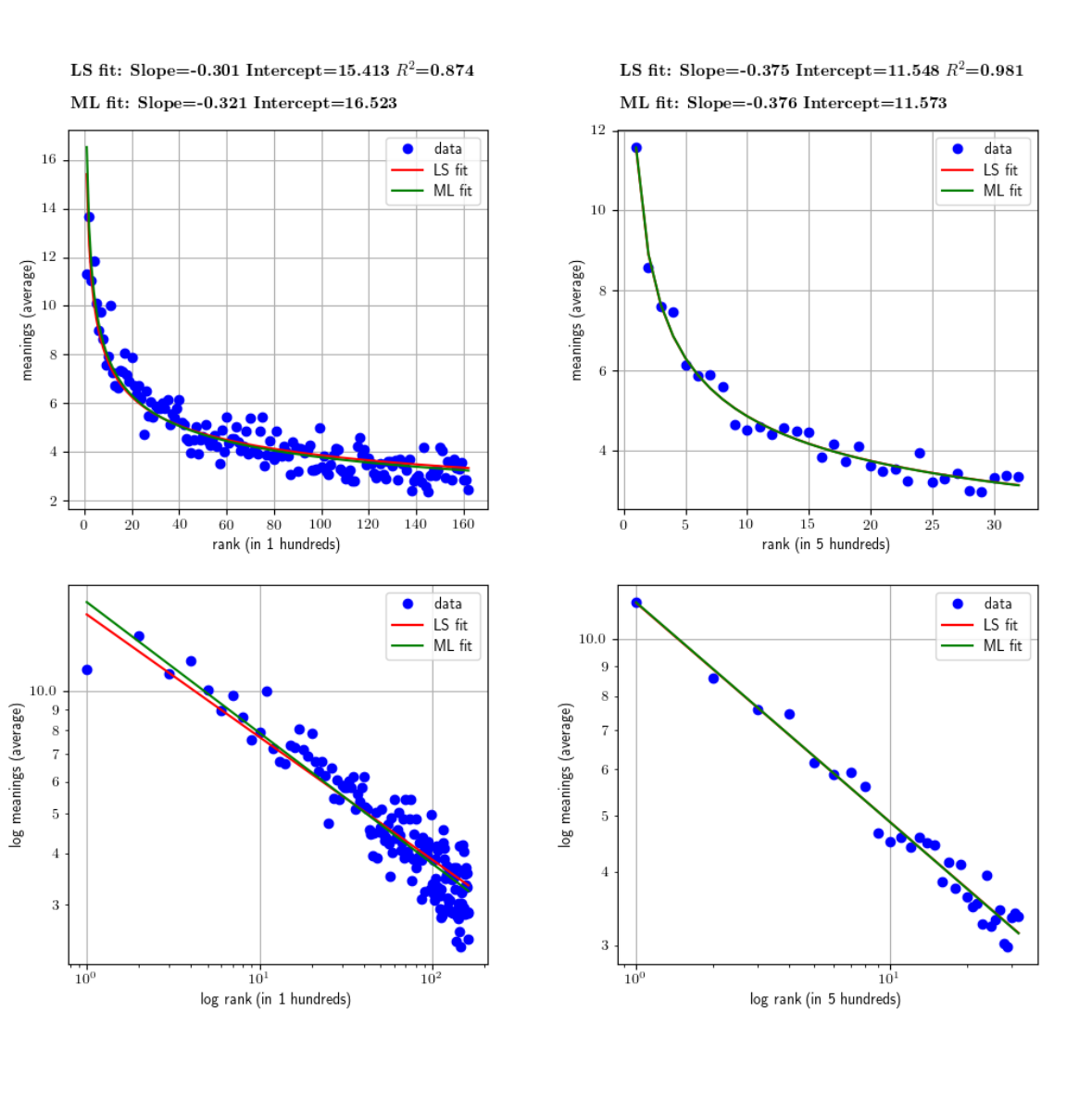}
\else
	\begin{minipage}{0.49\linewidth}
	\centering
	\includegraphics[width=\linewidth]{files/eng_freq_childes_tot_ADU_1}
	\end{minipage}
	\begin{minipage}{0.49\linewidth}
	\centering
	\includegraphics[width=\linewidth]{files/eng_freq_childes_tot_ADU_5}
	\end{minipage}
\fi
\caption{Average number of meanings per word (on the ordinate) 
for each successive set of $\lambda$ words on the abscissa: 
the 1st set has rank 1, the 2nd set has rank 2, ... etc. 
($\lambda$ is the number of words per bin to reduce). 
The true data points (blue circles) are compared against 
the power-law models fitted using non-linear least squares 
(LS; red line) and maximum likelihood (ML; green line). }
Top: normal scale. Bottom: log-log scale. 
Left: $\lambda = 100$. Right: $\lambda = 500$.
Data set: English words produced by adults, using the CHILDES frequency.
\label{fig:eng_freq_childes_tot_ADU}
\end{figure}

\begin{figure}[htp]
\centering
\ifarxiv
	\includegraphics[width=\textwidth]{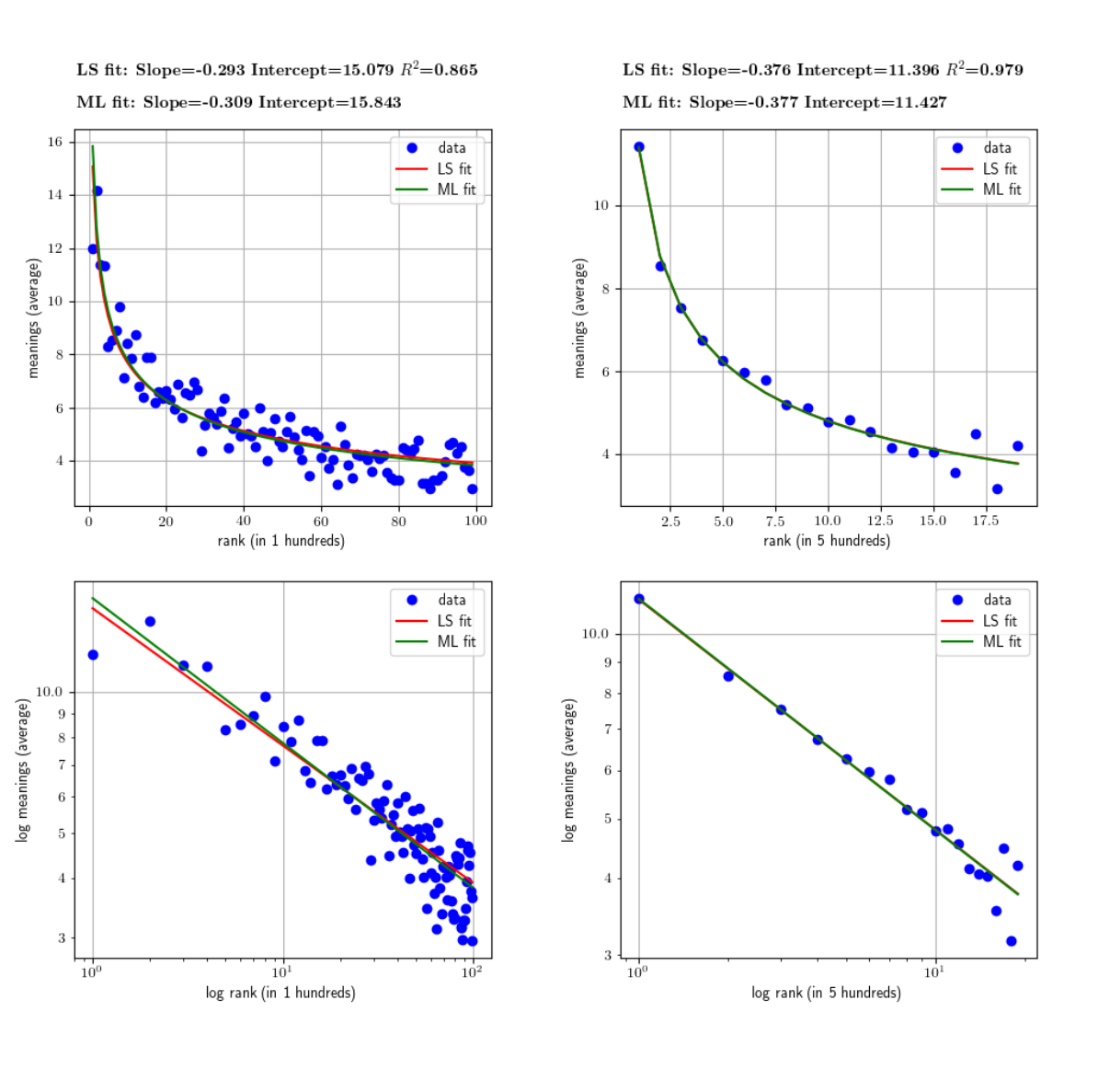}
\else
	\begin{minipage}{0.49\linewidth}
	\centering
	\includegraphics[width=\linewidth]{files/eng_freq_childes_tot_CHI_1}
	\end{minipage}
	\begin{minipage}{0.49\linewidth}
	\centering
	\includegraphics[width=\linewidth]{files/eng_freq_childes_tot_CHI_5}
	\end{minipage}
\fi
\caption{Same format as in Fig. \ref{fig:eng_freq_childes_tot_ADU} for English words produced by children. }
\label{fig:eng_freq_childes_tot_CHI}
\end{figure}

Figures \ref{fig:eng_freq_childes_tot_ADU} and \ref{fig:eng_freq_childes_tot_CHI}
show two examples of these plots taking as input
English words produced by adults and by children, respectively.
Frequencies have been obtained from CHILDES for both data sets.
For estimating the values of the parameters, the slope and
the Y-intercept of the best regression line, we have used two different methods:
non-linear least squares and maximum likelihood \cite{Seal1952}
on the original curve (in normal scale).
The values shown at the top of each figure correspond
to the parameters of the fitting in log-log scale,
that define the regression line.

Tables \ref{table:eng_freq_childes_tot_ADU_CHI_1} and \ref{table:eng_freq_childes_tot_ADU_CHI_5}
summarize the analyses performed over the whole data sets, that is,
for the three languages (English, Dutch and Spanish) and for the two roles
(adults and children).
Table \ref{table:eng_freq_childes_tot_ADU_CHI_1} corresponds
to groupings of 100 words (the abscissa of a point
represents 100 words and the value of its ordinate
is the average number of meanings per word in this hundred).
Table \ref{table:eng_freq_childes_tot_ADU_CHI_5} corresponds to groupings of 500 words.

\begin{table}[htp]
\centering
\scriptsize
\begin{tabular}{|c|c|c|c|c|c|c|}
\hline
\multirow{2}{*}{\textbf{Language / role}} & \multirow{2}{*}{\textbf{N}} & \multicolumn{3}{c|}{\textbf{Least squares}} & \multicolumn{2}{c|}{\textbf{Maximum likelihood}} \\
\cline{3-7}& & \textbf{slope} & \textbf{intercept}  & \textbf{R-squared} & \textbf{slope} & \textbf{intercept} \\
\hline
English/Adults & 162& \num{-0.300713604619} & \num{15.4125589744} & \num{0.874491051461} & \num{-0.320595724261} & \num{16.5225950173}\\
\hline
English/Children & 99& \num{-0.292547702747} & \num{15.0789884415} & \num{0.865110480709} & \num{-0.308511004682} & \num{15.8425467904}\\
\hline
Dutch/Adults & 52& \num{-0.161499220636} & \num{3.7226241012} & \num{0.742566141323} & \num{-0.168828345124} & \num{3.79931306942}\\
\hline
Dutch/Children & 26& \num{-0.170075461562} & \num{3.54879909075} & \num{0.776658895095} & \num{-0.175786930551} & \num{3.59282215562}\\
\hline
Spanish/Adults & 32& \num{-0.177085869014} & \num{6.2247369614} & \num{0.630504293523} & \num{-0.185885234788} & \num{6.35345064714}\\
\hline
Spanish/Children & 35& \num{-0.173148259615} & \num{6.02538654433} & \num{0.590873160994} & \num{-0.184443434924} & \num{6.19159498917}\\
\hline
\end{tabular}
\caption{Estimated values for the parameters of Zipf's law of meaning distribution by Least squares and by Maximum Likelihood. $N$ is the number of data points after grouping the words into groups of 1 hundred.}
\label{table:eng_freq_childes_tot_ADU_CHI_1}
\end{table}

\begin{table}[htp]
\centering
\scriptsize
\begin{tabular}{|c|c|c|c|c|c|c|}
\hline
\multirow{2}{*}{\textbf{Language / role}} & \multirow{2}{*}{\textbf{N}} & \multicolumn{3}{c|}{\textbf{Least squares}} & \multicolumn{2}{c|}{\textbf{Maximum likelihood}} \\
\cline{3-7}& & \textbf{slope} & \textbf{intercept}  & \textbf{R-squared} & \textbf{slope} & \textbf{intercept} \\
\hline
English/Adults & 32& \num{-0.37509334841} & \num{11.5483111649} & \num{0.981206724774} & \num{-0.376157575023} & \num{11.5731385349}\\
\hline
English/Children & 19& \num{-0.375579832513} & \num{11.3958343787} & \num{0.97895749176} & \num{-0.377237794711} & \num{11.4271123849}\\
\hline
Dutch/Adults & 10& \num{-0.224110735057} & \num{3.22339015353} & \num{0.974114193524} & \num{-0.226125578935} & \num{3.2320618777}\\
\hline
Dutch/Children & 5& \num{-0.252770880774} & \num{3.04830820895} & \num{0.99337183105} & \num{-0.253818755465} & \num{3.05095879987}\\
\hline
Spanish/Adults & 6& \num{-0.264532720579} & \num{5.36185280694} & \num{0.923829331367} & \num{-0.262683449737} & \num{5.35255007206}\\
\hline
Spanish/Children & 7& \num{-0.271026380887} & \num{5.28879665705} & \num{0.947752552154} & \num{-0.27199532053} & \num{5.29415020812}\\
\hline
\end{tabular}
\caption{Estimated values for the parameters of Zipf's law of meaning distribution (slope and intercept) by Least squares and by Maximum Likelihood. $N$ is the number of data points after grouping the words into groups of 5 hundreds.}
\label{table:eng_freq_childes_tot_ADU_CHI_5}
\end{table}

As shown in Tables \ref{table:eng_freq_childes_tot_ADU_CHI_1} and \ref{table:eng_freq_childes_tot_ADU_CHI_5}, as groupings become larger, the slopes of the regression
lines become closer to the expected value of $-0.5$ \cite{Zipf1949a, Ilgen2007a}.
We want to note here that, especially in the case of English,
groupings of 1000 words (as in Zipf's work \cite{Zipf1949a})
yield slopes of exactly $-0.5$.

These results suggest that using a similar methodology in \cite{Zipf1949a},
the analyzed data sets confirm Zipf's meaning-frequency law.
However, in this paper we want to present, an alternative way of confirming this law
(as well as the Zipf's law of abbreviation) by means of a correlation analysis.

The dependence of the exponent on bin size is a challenge for research on the law of meaning distribution. Paradoxically, when using the same bin size as Zipf did ($\lambda = 1000$), we get the number of non-empty bins reduced to a few for Spanish and Dutch (this is the reason why we exclude that bin size in the tables above). We can reduce the bin size to maximize the number of languages used but then the exponents deviate from the originally reported by Zipf in English. One also needs to control for the kind of binning. Logarithmic binning is often used when investigating power law relationships
\cite{Corral2006a}. In addition, we are fitting Eq. \ref{law_of_meaning_distribution_equation} and estimating $\gamma$ assuming that a power law holds. The validity of such assumption must be tested. The plots above suggest a deviation from the power law for high ranks.

Finally, we need to consider the role of the data source in the emergence of the power law. For instance, word frequency could be estimated from Wikipedia entries (see Section \ref{sec:methods} for such a possibility). The magnitude of the whole challenge can easily be reduced with a non-parametric approach based on a correlation analysis that does not involve neither any kind of binning nor assuming an exact model (an equation), that may not be generally valid. So, after revisiting the classical Zipfian approach to the meaning-frequency law \cite{Zipf1949a, zipf45meaning}, next sections develop our proposal of a correlation analysis between frequency, meaning and word length.

%%%%%%%%%%%%%%%%%%%%%%%%%%%%%%%%%%%%%%%%%%%%%%%%%%%%%%%%%%%%%%%%%%%%%%%%%%%
%
\section{Materials}
\label{materials}

In this section we describe the different corpora and tools that
have been used in this paper.
We first describe the WordNet database
which has been used to compute the polysemy measure.
We also describe the tools used to convert text to phonetic transcription
and to perform syllabic segmentation: CELEX database and SAGA.
Finally, we describe the two different sources for calculating the frequency of words that are analyzed in this
paper: CHILDES database and Wikipedia as reference corpora.

\subsection{Open Multilingual Wordnet}
\label{wordnet}

%One way to obtain the polysemy of a word in English is querying the lexical
%database WordNet \cite{Miller1995a, FellBaum1998a}.

The WordNet database can be seen as a set of senses (also called \emph{synsets})
and relationships among them, where
a synset is the representation of an abstract meaning,
and it is defined as a
set of words having (at least) the meaning that the synset stands for.
%Apart from this pair of sets, a relationship between both is also contained.
%As an example in WordNet, the word \emph{book} is related (among others) with the synset
%that represents \textit{a written work or composition that has been published},
%with the synset that represents \textit{a written version of a play or other
%dramatic composition; used in preparing for a performance} or the synset
%that represents \textit{to arrange for and reserve (something for someone else) in advance}.
Each pair word-synset is also related to a syntactic category.
For instance, the pair \emph{book} and the synset
\textit{a written work or composition that has been published}
are related to the category \emph{noun}, whereas the
pair \emph{book} and synset
\textit{to arrange for and reserve (something for someone else) in advance}
are related to the category \textit{verb}.

%\textcolor{red}{NEUS, BERNARDINO: En comptes de wordnets, fem servir WordNet database. Tambe unifiquem notacio: Wordnet -> WordNet, si cal}

Open Multilingual WordNet \cite{Bond:Foster:2013} gives access to open wordnets in several languages. In this paper we use the Princeton WordNet for English, the Open Dutch WordNet for Dutch and the Multilingual Central Repository for Spanish.

%version for English, Dutch and Spanish, provided by the NLTK platform (Natural Language Toolkit)\footnote{Freely available at \url{http://www.nltk.org}}.
%that are freely available for download at
%\url{http://compling.hss.ntu.edu.sg/omw/}.
Since each WordNet has been made by many different projects, they all vary notably in size and coverage. Table \ref{tab:statsWN} shows some statistics for every WordNet used in this paper.
%: \emph{number of synsets}, \emph{number of words} and \emph{core}. \emph{Core} refers to the percentage of synsets covered from the semi-automatically compiled list of 5000 ``core'' word senses in Princeton WordNet (approximately the 5000 most frequently used word senses).
%\textcolor{green}{FORA per redundant i poc informatiu:
%These statistical values reveal differences of coverage,
%size and senses per word distribution among different WordNet databases.}

%\textcolor{green}{BERNARDINO: aixo es cert? ens sortien 
%paraules que no eren
%noms verbs adjectius o adverbis. NEUS responent: si, 
%wn conte nomes aquestes 4 categories sintactiques, 
%perque les preposicions, articles, determinants 
%no tenen contingut semantic... PERO podem trobar, p.e., 
%preposicions lligades a verbs per interpretar el seu sentit: 
%climb up, climb down; tambe podem trobar a wn 'a' 
%pensant que es un article i resulta que te mil 
%sentits pero no com article sino: la vitamina a, 
%la unitat angstrom (A), l'adenina (A), l'ampere (A)...}
WordNet databases contain only
four main syntactic categories: nouns, verbs,
adjectives and adverbs.
Words of other syntactic 
categories are not present in these
databases 
(for instance, in English the article 
\emph{the} or the preposition \emph{for}).
However, some words which should be considered as functional words,
have been included in our analyses, because they can also
be considered as content words 
(i.e. in English, the determiner \textit{a} can also be a noun as
in \textit{Letter A} or \textit{Vitamin A}).

\statsWN

\subsection{CELEX database}
\label{sec:celex}

CELEX \cite{CELEX} is the Dutch Centre for Lexical Information at the Max Planck Institute for Psycholinguistics.
CELEX database comprises three different searchable lexical databases, Dutch, English and German. %In this paper we only use the information of Dutch and English databases.
%was used to have a reference word frequency
%that is external to the corpora that we have processed. The database
%is based on corpora in Dutch, English and German.
The lexical data contained in each database is divided into five categories:
orthography,
phonology,
morphology,
syntax (word class) and
word frequency.
% based on recent and representative text corpora.   % nomes per word-frequency i no es gaire recent

We use CELEX database to obtain the phonetic transcription and syllabic segmentation of Dutch and English speech transcripts. Using WebCelex\footnote{\url{http://celex.mpi.nl/}} we have created two lexicons, one for
English and another for Dutch, by selecting from the English Wordforms and Dutch Wordforms databases the following items:
\emph{Word} (from the orthography category), and \emph{PhonSAM} and \emph{PhonSylSAM} (from the phonology category).
The format of the phonetic transcription is SAMPA charset (Speech Assessment Methods Phonetic Alphabet\footnote{\url{http://www.phon.ucl.ac.uk/home/sampa/index.html}}).

\subsection{SAGA}
\label{sec:saga}

SAGA is an automatic tool for phonetic transcription in Spanish, considering its multiple dialectal variants. The phonetic description is made in terms of the SAMPA alphabet. The tool is able to split the words into syllables and mark the prosodic stress.

SAGA is able to perform different kinds of transcriptions depending on the output settings (phonemes, semi-phonemes, syllables, semi-syllables). In addition, even Spanish has a mostly phonetic writing, there are some exceptions to the general phonetic rules as for example foreign words, archaic language or dialectal variants. To deal with these cases, SAGA contains dictionaries that can be modified to customize the phonetic transcriptions as desired.

We have used SAGA for Spanish conversations to perform both phonetic transcription and syllabic segmentation. This application is distributed under the terms of the GNU General Public License\footnote{Freely available at \url{http://www.talp.upc.edu/index.php/technology/tools/signal-processing-tools/81-saga}}.
%It is distributed under the GNU General Public License (http://www.gnu.org/licenses/gpl-3.0.txt).

\subsection{Wikipedia}

Wikipedia is a free online encyclopedia built collaboratively and hosted by the non-profit Wikimedia Foundation. It exists in 295 languages, from which currently there is a total of 284 active ones, with the number of pages ranging from more than 5 million articles (English) to a few hundred articles (Zulu, Romani, Greenlandic\ldots)\footnote{\url{https://en.wikipedia.org/wiki/List_of_Wikipedias}}.

Wikipedia includes articles that span across many topics and it is updated with constant contributions. Thus, it turns out to be a useful resource as a reference corpus for getting word frequencies. Since we use two different sources for estimating word frequencies, we can compare the results obtained by using a general corpus (Wikipedia) with the use of a simpler one (CHILDES).

The contents of each Wikipedia can be downloaded and processed to calculate the frequency of every word that appears in Wikipedia \cite{GREFENSTETTE16.624}. We have downloaded from
Gregory Grefenstette webpage the lexicons with the frequencies extracted from Wikipedia for English, Dutch and Spanish
% Neus: trec el footnote perque l'enllac esta trencat
% \footnote{\url{https://www.lri.fr/~ggrefens/}}
\footnote{\url{http://web.archive.org/web/20170205022929/http://pages.saclay.inria.fr:80/gregory.grefenstette/}}.

\subsection{CHILDES database}
\label{sub:childes}

The CHILDES database
%\footnote{CHILDES database is freely available for download at
%\url{http://childes.psy.cmu.edu/data} (accessed 17 December 2012).}
\cite{MacWhinney2000a}
is a set of corpora of transcripts of
conversations between children and adults.
The corpora included in this database
are in different languages.
%, and contains conversations when the children
%were between 12 and 65 months old, approximately.

In this paper we have studied the conversations of 60 children in English, 73 children in Dutch and 490 children in Spanish.
Detailed information on these conversations can be found in Table \ref{tab:childes_british} for British English,
Table \ref{tab:childes_american} for American English, Table \ref{tab:childes_dutch} for Dutch and \ref{tab:childes_spanish} for Spanish
in \ref{sec:appendixchildes}.
%\cite{Baixeries2012c}).
%, which will be referred to as
%\textbf{recording sessions} along this paper.
%We have processed every conversation of the selected corpora of CHILDES.
For each spoken word of these conversations the following values are given: CHILDES frequency (number of times this word appears in CHILDES, counted separately by children and adults),
Wikipedia frequency (number of times this word appears in Wikipedia),
number of synsets (according to the corresponding WordNet),
number of characters, number of phonemes and number of syllables.
%We have only taken into account the words that appear in WordNet, that is content words (nouns, verbs, adjectives and adverbs).
Table \ref{tab:analyzed-words} shows
the number of different types and tokens obtained from the selected corpora.
The number of analyzed tokens and types is smaller than
the number of tokens and types initially extracted from the conversations, because only those words that are present in the correspondent WordNet have been retained.
%\textcolor{red}{aquesta ultima frase no l'entenc -- resolt}

\analyzedWords

\section{Methods}
\label{sec:methods}

We now describe the different numerical and computational methods that have been used in this paper.

%In this paper we study the relationship between three variables
%that are related to every word: length, frequency and polysemy.

\subsection{Word Length Computation}

There are several types of units to measure word length among
which the most used are graphic and phonetic. Graphical units
are usually characters or letters. Phonetic units are phonemes,
syllables or sounds, and although they are highly correlated
with graphical units there can be differences depending on the language.

Here, we have considered three different units of measurement:
a graphical unit (number of characters) and two phonetic units
(number of phonemes and number of syllables). When dealing
with counting characters, numbers, blanks, separation
characters and the like have not been taken into consideration.
The resources used to obtain orthographic and phonetic information are described in
Section \ref{materials}.

\subsection{Frequency}

We have extracted word frequency values from two different sources. Thus, for each word that appears in
the selected conversations of CHILDES, we obtain:

\begin{itemize}
\item \textbf{Wikipedia frequency}, the frequency that
the given word has in the Wikipedia dataset.

\item \textbf{CHILDES frequency}, the frequency that the given word has in
CHILDES according to the speaker's role: children or adults
(comprising mothers, fathers and investigators).
For example, for the word \textit{book}
two different frequencies are given:
the number of times this word appears uttered by
children and uttered by adults, respectively.
\end{itemize}

%SemCor frequency can only be analyzed in the SemCor corpus,
%whereas CELEX and CHILDES frequencies are only analyzed
%in the CHILDES corpora.

\subsection{Polysemy}

From linguistics, polysemous words are words that have more
than one meaning. Linguists distinguish between words with multiple meanings,
where the meanings are unrelated (called homonyms), and words with multiple senses,
where the senses are related. An example of the former is the word \emph{bank},
having unrelated meanings such as \texttt{a sloping land} or
\texttt{a financial institution}, whereas an example of
the latter is \emph{honey}, having related senses such as
\texttt{sweet yellow liquid produced by bees} or \texttt{a beloved person}.

We have calculated the polysemy of a word as the number of
different meanings provided by the WordNet database of its
corresponding language. In WordNet, the different senses
of a polysemous word are assigned to different \emph{synsets}.
Then, we have considered the number of synsets a word belongs
to as the number of meanings it has. This count is what
we call WordNet polysemy. We assume that the polysemy measure
provided by WordNet does not differentiate between
polysemy classes mentioned above.

We are aware that using the WordNet polysemy measure in the CHILDES corpora induces a bias.
First, because we are assuming that the same meanings that
are used in written text are also used in spoken language.
Second, because we are using all possible meanings of a word.
An alternative would have been to tag manually all corpora
(which is currently an unavailable option) or to use an automatic tagger.
But also in this latter case, the possibility of biases or errors would be present.

%In this paper, we use Semcor polysemy in the Semcor corpus,
%and both Semcor and Wordnet polysemy for the Childes corpus.

\subsection{Statistical Methods}

In the present work we have studied the relationship between
(1) frequency and polysemy and (2) frequency and length.
For the three variables, frequency, polysemy and word length,
we have used different sources yielding us many combinations for evaluation.

%For the CHILDES corpora, the availability of
%different sources for frequency and polysemy yields the following combinations:

For each language selected in the CHILDES corpora, we have calculated correlations between:

\begin{enumerate}
\item CHILDES frequency and WordNet polysemy
\item CHILDES frequency and number of characters
\item CHILDES frequency and number of phonemes
\item CHILDES frequency and number of syllables
\item Wikipedia frequency and WordNet polysemy
\item Wikipedia frequency and number of characters
\item Wikipedia frequency and number of phonemes
\item Wikipedia frequency and number of syllables
\end{enumerate}

For each combination of two variables, we compute:

\begin{enumerate}
\item \textbf{Correlation test}.
Pearson, Spearman and Kendall two-sided correlation tests \cite{Conover1999a},
using the \texttt{cor.test} standardized R function.
The traditional Pearson correlation is a measure of linear dependency
while Spearman and Kendall correlations are to capture non-linear dependencies
\cite{Gibbons2010a,Embrecths2002a}.
%\textcolor{red}{
%Pearson's method supposes input vectors approximately normally distributed while
%Spearman's is a non-parametric test
%that does not require vectors being approximately normally distributed \cite{Baayen2007a}.
%Kendall's tau is more robust to extreme observations and to non-linearity
%compared with the standard Pearson product-moment correlation \cite{Newson2002}.
%}

\item \textbf{Plot}, in logarithmic scale,
that also shows the density of points.

\item \textbf{Nonparametric regression}, to obtain a smoothed curve for the cloud of points defined by the two variables. The smoothed curve is calculated using the \texttt{locpoly} standardized R function and added to the previous plot.

\item \textbf{Probability density function} using local polynomials. Proportional density function is calculated using the \texttt{locpoly} standardized R function and added to the previous plot.
%using the \texttt{locpoly} standarized R function, which has been overlapped in the previous plot.
\end{enumerate}

On top of the correlation analysis we build another analysis where we compare pairs of correlations that have a variable in common. Our goal is to determine if the unit of measurement has some effect on the strength of a linguistic law. When we investigate the law of abbreviation (the correlation between the frequency of a word and its length), we keep the source used to estimate frequency while we vary the way length is measured: number of characters, number of phonemes or number of syllables.
 %or time (duration).
In particular, we determine if the difference between
two dependent correlations sharing one variable is significant.

Suppose that we have two different length measures
$L_1, L_2$ (which can be the number of characters, phonemes or syllables)
and one frequency measure $F$ (which can be the CHILDES or Wikipedia frequency).
Suppose that the correlation between $F$ and $L_1$ is $r(F, L_1)$
and the correlation between $F$ and $L_2$ is $r(F, L_2)$,
and that both correlations are negative. % and $|r(F, L_1)| < |r(F, L_2)|$.
To determine if one of those correlations
is significantly stronger that the other,
%if the difference between $r(F, L_1)$ and $r(F, L_2)$ is significant,
we apply a two-tailed Steiger's test \cite{Steiger1980a}
(we use the \texttt{r.test} standardized R function).
If the p-value is below the significance level
and $|r(F, L_1)| < |r(F, L_2)|$,
we can conclude that $L_2$ is more correlated to $F$ than $L_1$.
Else, if $|r(F, L_1)| > |r(F, L_2)|$,
we can conclude that $L_1$ is more correlated to $F$ than $L_2$.
%$L_2$ is more correlated to
%$F$ than $L_1$.
Otherwise, if the test is non significant,
we cannot conclude that one correlation is stronger than the other.

We note that the \texttt{r.test} standardized R function requires
a single sample size as a parameter. For this reason,
before performing the test,
in order to compute $r(F, L_1)$ and $r(F, L_2)$,
we have selected from the dataset,
those records that have a valid value on \textbf{all} three variables
($F, L_1, L_2$).
However, when we compute a correlation test 
two single variables $F$ and $L_1$ (or $L_2$),
we select all those records that have a valid value
in \textbf{both} $F$ and $L_1$ ($L_2$),
but not necessarily in all three of them.
Therefore, the value computed for $r(F, L_1)$ and $r(F, L_2)$
in the Steiger's test may yield a somehow different value from
that in a single correlation test because of this constraint
(inherent to the Steiger's test).

As the theory of Steiger's test is defined on Pearson correlation,
this higher level analysis is performed only on Pearson correlations
and Spearman correlations (Spearman correlation
is a Pearson correlation on rank transformation of
the random variables \cite{Conover1999a}).
As far as we know, it is not warranted
that the Steiger's tests can be applied on Kendall correlations. % ask Chris Kello about versions for Kendall ???

%TONI: Pujat a dalt, en verd, la justificació de les tres correlacions que hi havia aquí, més a prop d'on es presenta.

%\textcolor{green}{
%FORA, és reduntant amb material de la secció del test d'steiger. Ho he ajuntat tot aquí.
%More formally, suppose that $f(x,y)$ measures the correlation between two variables $x$ and $y$. We determine if $|f(x,y) - f(x,z)|$ is statistically large with the help of a two-tailed Steiger's test \cite{Steiger1980a} (we use the \texttt{r.test} standardized R function). $x$ is word frequency and $y$ could be word length in terms of character numbers and $z$ could be word length in terms of syllable numbers.
%As the theory of Steiger's test is defined on Pearson correlation, this higher level analysis is performed only on Pearson correlations and Spearman correlations (Spearman correlation is a Pearson correlation on rank transformation of the random variables \cite{Conover1999a}). % ask Chris Kello about versions for Kendall ???
%}

We use three different measures of correlations for the following reasons.
Pearson correlation is included for its popularity and simplicity.
Spearman and Kendall correlation are included for
their capacity to capture non-linear dependencies.
Spearman is needed for the Steiger's tests (see Section \ref{results})
and Kendall correlation allows one to interpret the strength
of a correlation based on the number of ties
(this will be shown in Section \ref{ties_subsection}).

We assume a significance level of 0.05 in all tests.

We remark that the analysis for the CHILDES corpora has been segmented
into two roles: children and adults.
%by speakers' roles.

%%%%%%%%%%%%%%%%%%%%%%%%%%%%%%%%%%%%%%%%%%%%%%%%%%%%%%%%%%%%%%%%%%%%%%%%%%%
%
\section{Results}
\label{results}

We describe the results that have been obtained in three different languages 
(English, Dutch and Spanish) from the analysis of the relationship between:

\begin{enumerate}
\item Frequency and polysemy (meaning-frequency law).
\item Frequency and word length (Zipf's law of abbreviation).
\end{enumerate}

We use two different measures for frequency (CHILDES frequency
and Wikipedia frequency),
one measure for polysemy (WordNet polysemy)
and three measures for word length
(number of characters, phonemes and syllables)
as previously explained in Section \ref{sec:methods}.

Here we present the results in two formats:

\begin{enumerate}
\item A table that contains the results of
a correlation test between a frequency measure
versus the following measures:
WordNet polysemy,
number of characters, number of phonemes and
number of syllables.
Each table shows the results of
three (Pearson, Spearman and Kendall) correlation tests.
For each language we have produced two tables:
one table where the frequency measure is the CHILDES frequency,
and another table where the frequency measure is the Wikipedia frequency.

\item A plot for each pair of variables that have been analyzed in the
previous tables along with a nonparametric regression and
a probability density function
(see Section \ref{sec:methods} for details).
\end{enumerate}

We also present the results of the Steiger's test that shed light 
on which of the three different length measures
exhibits a stronger correlation with frequency.
Finally, we include a subsection in which we examine the
impact of ties in our analyses.

%\taulaTypesTok		% typestokes

	\subsection{Frequency versus Polysemy}

We now describe the results that analyze the relationship between
frequency and polysemy.
We remind the reader that we compare the two sources of frequency
(CHILDES and Wikipedia) with WordNet polysemy.

In English, all correlations are significant and positive
(Tables \ref{corENGchi} and \ref{corENGwik} in the \ref{sec:appendixcorrelations}).
%\corrEngCHILDES		% corENGchi
%\corrEngWiki		% corENGwik
In Dutch (Tables \ref{corDUTchi} and \ref{corDUTwik} 
in the \ref{sec:appendixcorrelations})
and Spanish (Tables \ref{corSPAchi} and \ref{corSPAwik} 
in the \ref{sec:appendixcorrelations}),
Spearman and Kendall correlations are significant and positive
and the Pearson correlations are non-significant 
with a correlation value close to zero.

% There is an exception to this pattern: the value of the Pearson correlation 
% between CHILDES frequency and Wordnet polysemy is negative in a subset of 
% the adults production (fathers and investigators 
% \textcolor{red}{Toni comment: ***SI POSEM AIXÒ, com alguns comentaris 
% següents on parlem de pares, mares, etc., LLAVORS LES TAULES DE DADES 
% SEPARADES HAN D'ANAR COM A SUPLEMENTARY INFORMATION O ANNEX***}), 
% but these values are very close to zero.
% In fact, the average value for all roles is $-0.002$ with a very low standard deviation: 
% $0.017$. A similar case is found when correlating Wikipedia frequency and 
% Wordnet polysemy: fathers ($-0.01$) have a slight negative Pearson correlation value, 
% but the average is, again, very small: $0$ with a standard deviation of $0.007$.

%\corrNldCHILDES		% corDUTchi
%\corrNldWiki		% corDUTwik

%In Spanish (Tables \ref{corSPAchi} and \ref{corSPAwik} 
%in the \ref{sec:appendixcorrelations}),
%all correlations are significant and positive,
%with the exception of the Pearson correlation between
%CHILDES frequency and Wordnet polysemy,
%which is almost zero.
%% , and, in the few cases that they are negative, their values are very close to zero
%% (children and subset of fathers in the Pearson correlation).
%This is probably due to lower capacity of that statistic 
%to detect non-linear dependencies compared to the other statistics.

%\corrSpaCHILDES		% corSPAchi
%\corrSpaWiki		% corSPAchi

A visual inspection of the graphics 
in Figure \ref{fig:freqchildes_vs_nsynwordnet}
in the \ref{sec:grafiques}
confirms the patterns that we have examined previously.
When CHILDES frequency is analyzed,
we see in English that, in all cases, the nonparametric regressions
show a positive slope in the area where most of the
points are concentrated. Where this concentration decreases,
so does the nonparametric regression.
In both Dutch and Spanish, the regression shows a similar pattern,
but the increase is not as strong as in English.
But, as in English, the regression decays significantly when it
reaches the area with a smaller density of points.

When Wikipedia frequency is analyzed
(Figure \ref{fig:freqwiki_vs_nsynwordnet}
in the \ref{sec:grafiques}),
we can observe two facts:
points are distributed in a more compact way and
the nonparametric regression has a steeper slope in English,
and, to a lesser degree, in Dutch and Spanish.

%\figura{freqchildes}{nsynwordnet}{WordNet polysemy versus CHILDES frequency 
%in double logarithmic scale. The color indicates the density of points: dark green is the
%highest possible density. A smoothed curve (blue) and a curve proportional 
%to the probability density of values of the $x$-axis (red) is also shown.}	
%% fig:freqchildes_vs_nsynwordnet
%\figura{freqwiki}{nsynwordnet}{WordNet polysemy versus Wikipedia frequency. 
%The format is the same as in Fig. \ref{fig:freqchildes_vs_nsynwordnet}. }	
%% fig:freqwiki_vs_nsynwordnet

To sum up, we can say that, in a vast majority of cases, 
the values of the significant correlations are always positive.
Correlations are significant in English in all cases,
and for all non-linear correlation measures
%in majority of cases 
in Dutch and Spanish.
% The few cases in which the correlation values have a negative value can
% be seen as exceptions, since these values are very close to zero,
% as well as the average the correlation values of all roles taken together.
% It is also interesting to note that the correlations that are not significant
% are positive in all cases.

%\textcolor{red}{***Toni comment: Jo afegiria, com us deia, les dades separades per rols (ja que les tenim, com annex final. Llavors es %poden citar i esborrar el meu comentari anterior. ELS PEUS DE FIGURA HAN DE SER AUTOEXPLICATIUS.***}

\subsection{Frequency versus Length}

\label{frequency_versus_length_subsection}

The analysis of the two measures of frequency
versus the three measures of length are in
Tables \ref{corENGchi} and \ref{corENGwik} in English,
Tables \ref{corDUTchi} and \ref{corDUTwik} in Dutch,
and
Tables \ref{corSPAchi} and \ref{corSPAwik} in Spanish in the  \ref{sec:appendixcorrelations}.
In this case, the results show a more
compact behavior, since
all correlations are significant and negative both for children and for adults.
%with only one single exception:
%the correlation between Wikipedia frequency and
%all three measures of length in Spanish
%are non-significant (with a very few exceptions)
%and their values are positive and very close to zero
%(not greater than $0.04$).

As for the nonparametric regression
in Figures \ref{fig:freqchildes_vs_nlletres},
\ref{fig:freqwiki_vs_nlletres},
\ref{fig:freqchildes_vs_numphon},
\ref{fig:freqwiki_vs_numphon},
\ref{fig:freqchildes_vs_numsylab} and
\ref{fig:freqwiki_vs_numsylab},
we have that the
results are consistent with these previous patterns:
in all cases, the regression shows a negative slope.

\subsection{Steiger's test}

\label{Steiger_test_subsection}

We have seen in the previous section that
all three measures of length exhibit a negative correlation
with respect to frequency.
We now turn into the question of deciding
which of these measures holds the strongest correlation with frequency
by means of a Steiger's test
(see Section \ref{sec:methods} for details about how this test
has been computed).

%\textcolor{green}{
%FORA: ho poso a la secció de mètodes.
%Suppose that we have two different length measures
%$L_1, L_2$ (which can be the number of characters, phonemes or syllables)
%and one frequency measure $F$ (which can be the CHILDES or Wikipedia frequency).
%Suppose that the correlation between $F$ and $L_1$ is $r(F, L_1)$
%and the correlation between $F$ and $L_2$ is $r(F, L_2)$
%and that both correlations are negative and $|r(F, L_1)| < |r(F, L_2)|$.
%To determine if the difference between $r(F, L_1)$ and $r(F, L_2)$ is significant, we apply a Steiger's test. If the test indicates that the difference is significant then we can conclude that $L_2$ is more correlated to
%$F$ than $L_1$. Otherwise, we cannot conclude that one is stronger than the other.
%}

In \ref{sec:appendixsteiger} we present,
for each pair of length measures 
(which can be the number of characters, phonemes or syllables)
a table that displays the analytical
results of the Steiger's test
(the \textbf{t} and \textbf{p-value})
with respect to the two different sources of frequency (CHILDES or Wikipedia).
Table \ref{nlletresnphon} shows the results for the Steiger's test
between variables number of characters and number of phonemes,
Table \ref{nlletresnsyllab} shows the results 
between number of characters and number of syllables
and
Table \ref{nphonnsyllab} shows the results between variables
number of phonemes and number of syllables.

We also provide a more compact way of seeing the results contained
in those tables in a set of three different tables, one for
each language:
English in Table \ref{ntaulaSteigerENG},
Dutch in Table \ref{ntaulaSteigerNLD}
and Spanish in Table \ref{ntaulaSteigerSPA}.
In these tables, for each analyzed language, 
we list all possible combinations of role,
frequency measure and correlation type and,
for each combination, we display the order relationship 
on the strength of the correlation
for each pair of length variables.
In the column \textit{Char. vs Phon.} we show the relation between
number of characters and number of phonemes, 
in the column \textit{Phon. vs Syllables} we show the relation between
number of phonemes and number of syllables, and
in the column \textit{Phon. vs Syllables} we show the relation between
number of phonemes and number of syllables.
For each pair, the relation may be $>^*$ or $<^*$ if the Steiger's test
has determined that the difference of the correlations is significant.
If the test is not significant, then, the relation may be $>$ or $<$.
Since the Kendall correlation test is not analyzed with
the Steiger's test (see Section \ref{sec:methods}),
we have adopted the convention of assuming that this
test is always non-significant.

Tables \ref{ntaulaSteigerENG}, \ref{ntaulaSteigerNLD},
and \ref{ntaulaSteigerSPA}, provide the following results:
in English most of the tests are significant, and the pattern
that emerges more frequently is that of 
$Char. >^* Phon. >^* Syllables$, this is, that 
the number of characters is the length measure most correlated 
to a frequency measure, 
followed by the number of phonemes and, then, 
by the number of syllables.
In fact, the patterns $Char. >^* Syllables$
and $Phon. >^* Syllables$ appear in the vast
majority of cases.
In Dutch, we observe that little can be said about the
prominence of the number of phonemes or characters as
the most correlated length variable for lack of significance.
However, the two patterns $Char. >^* Syllables$
and $Phon. >^* Syllables$ (as in English)
appear in most of the cases.
As for Spanish, nothing can be said
because most of the relations are non-significant.

To sum up, the variables \textit{number of characters} and 
\textit{number of phonemes} show a stronger correlation
with respect to frequency than the number of syllables when the Steiger's test was significant.
The results also reveal (in English) a slightly stronger correlation between the number of characters
and frequency than between the number of phonemes and frequency.

\taulaSteigerENG
\taulaSteigerNLD
\taulaSteigerSPA

%\lletresphon 		% nlletresnphon
%\lletressyllab 		% nlletresnsyllab
%\phonlletres 		% nphonnsyllab

	\subsection{Proportion of ties}

\label{ties_subsection}

Here we aim at shedding some light on the weakness of the correlation between 
frequency and other variables. We focus on the Kendall $\tau$ correlation 
because it allows for a simple analysis of the influence of tied values.

The Kendall $\tau$ correlation is defined as \cite{Conover1999a}
\begin{equation}
\tau = \frac{N_c - N_d}{{n \choose 2}}
\label{Kendall_tau_correlation_equation}
\end{equation}
where $n$ is the sample size and $N_c$ and $N_d$ are, respectively, the number of concordant and discordant pairs in the sample.
We have that
\begin{equation*}
N_c + N_d + N_t = {n \choose 2}
\end{equation*}
where $N_t$ is the number of tied pairs (pairs that are neither concordant nor discordant).
Applying
\begin{equation*}
 N_d = {n \choose 2} - N_c - N_t
\end{equation*}
one can rewrite Eq. \ref{Kendall_tau_correlation_equation} equivalently as
\begin{equation}
\tau = -1 + \upsilon + \frac{2N_c}{{n \choose 2}}
\label{new_Kendall_tau_correlation_equation}
\end{equation}
where
\begin{equation*}
\upsilon = \frac{N_t}{{n \choose 2}}
\end{equation*}
is the proportion of tied pairs ($0 \leq \upsilon \leq 1$).
The fact that $N_c \geq 0$ allows one to see that
\begin{equation}
\tau \geq \upsilon - 1
\label{lower_bound_kendall_tau_correlation_equation}
\end{equation}
Put differently, the strongest negative Kendall $\tau$ that can be obtained is $\upsilon - 1$. The higher the number of ties, the weaker the maximum Kendall $\tau$ correlation that can be obtained.
Table \ref{table:ties} shows the percentage of ties, namely $100\upsilon$, of frequency (CHILDES and Wikipedia) versus polysemy and the measures of length for every language and role.

It is possible to derive lower bounds for the Spearman correlation $\rho$ from that of Kendall $\tau$ correlation. Knowing that \cite{Daniels1950a}
\begin{equation*}
\rho \geq \frac{1}{2} (3 \tau - 1)
\end{equation*}
and recalling Eq. \ref{lower_bound_kendall_tau_correlation_equation}, one obtains
\begin{equation}
\rho \geq \frac{3}{2} \upsilon - 2
\label{lower_bound1_spearman_rho_correlation_equation}
\end{equation}
Similarly, knowing that \cite{Durbin1951a}
\begin{equation*}
\rho \geq \frac{1}{2}(1+\tau)^2 - 1
\end{equation*}
one obtains
\begin{equation}
\rho \geq \frac{\upsilon^2}{2} - 1
\label{lower_bound2_spearman_rho_correlation_equation}
\end{equation}
Combining, Eqs. \ref{lower_bound1_spearman_rho_correlation_equation} and \ref{lower_bound2_spearman_rho_correlation_equation}, we get finally
\begin{equation}
\rho \geq max\left(\frac{3}{2} \upsilon - 1, \frac{\upsilon^2}{2}\right) - 1
\label{lower_bound_spearman_rho_correlation_equation}
\end{equation}
The lower bounds of $\rho$ above are likely to be looser than the original lower bound of $\tau$ because the former are derived from the latter.

\begin{table}[htb!]
\begin{center}
\resizebox{\columnwidth}{!}{
\begin{tabular}{|c|c|c|C{1.65cm}|C{1.65cm}|C{1.65cm}|C{1.65cm}|}
\hline
Language & Role & Frequency & Polysemy & Number of characters & Number of phonemes & Number of syllables\\
\hline
\hline
\multirow{4}{*}{English} & \multirow{2}{*}{Children} & CHILDES & 18.2\% & 20.9\% & 21.4\% & 35.4\% \\
\cline{3-7}
& & Wikipedia & 11.9\% & 14.4\% & 15.0\% & 29.9\% \\
\cline{2-7}
& \multirow{2}{*}{Adults} & CHILDES & 21.3\% & 21.3\% & 21.3\% & 33.9\% \\
\cline{3-7}
& & Wikipedia & 13.7\% & 13.2\% & 13.2\% & 26.9\% \\
\hline
\hline
\multirow{4}{*}{Dutch} & \multirow{2}{*}{Children} & CHILDES & 31.8\% & 19.5\% & 21.9\% & 38.0\% \\
\cline{3-7}
& & Wikipedia & 26.9\% & 13.1\% & 15.6\% & 33.0\% \\
\cline{2-7}
& \multirow{2}{*}{Adults} & CHILDES & 36.1\% & 21.8\% & 23.3\% & 37.0\% \\
\cline{3-7}
& & Wikipedia & 28.2\% & 11.1\% & 12.7\% & 28.4\% \\
\hline
\hline
\multirow{4}{*}{Spanish} & \multirow{2}{*}{Children} & CHILDES & 24.0\% & 21.4\% & 21.0\% & 38.7\% \\
\cline{3-7}
& & Wikipedia & 17.5\% & 14.3\% & 13.8\% & 33.0\% \\
\cline{2-7}
& \multirow{2}{*}{Adults} & CHILDES & 23.2\% & 21.5\% & 21.1\% & 38.3\% \\
\cline{3-7}
& & Wikipedia & 16.4\% & 14.1\% & 13.7\% & 32.3\% \\
\hline
\end{tabular}
}
\end{center}
\caption{Percentage of ties, $100\upsilon$, between frequency (CHILDES and Wikipedia) and WordNet polysemy, number of characters, number of phonemes and number of syllables for every language and role.}
\label{table:ties}
\end{table}

% \textcolor{green}{
% Cal revisar la discussio dins d'aquesta seccio.
% Punt clau: l'analisi dels empats ens ha de servir per interpretar les taules de correlacions o més aviat per interpretar els resultat del test de Steiger???
% }

In Sections \ref{frequency_versus_length_subsection} and \ref{Steiger_test_subsection}, we have shown a tendency of syllabic length to be the unit of length that is the most weakly correlated with frequency. This could be due to the higher proportion of ties of syllabic length ties in general (Table \ref{table:ties}), that reduces the potential strength of the correlation according to Eqs. \ref{lower_bound_kendall_tau_correlation_equation} and \ref{lower_bound_spearman_rho_correlation_equation}.

% \textcolor{green}{
% The values of $\upsilon$ between frequency and number of syllables are higher than other measures of length. That fact could affect to the Pearson and Spearman (precisament es de Kendall de qui podem (o creiem que podiem) fer els arguments mes forts ???) correlations within frequency and syllables that it is lower than other correlations of length measures in English and Dutch (see Tables \ref{corENGchi},  \ref{corENGwik}, \ref{corDUTchi} and \ref{corDUTwik}).
% }

% \textcolor{green}{Moreover in general the number of ties is higher in CHILDES frequency than Wikipedia frequency.
% }

%%%%%%%%%%%%%%%%%%%%%%%%%%%%%%%%%%%%%%%%%%%%%%%%%%%%%%%%%%%%%%%%%%%%%%%%%%%
%
\section{Discussion and Future Work}
\label{discussion}

In this paper, we have reviewed two linguistic laws that we owe to Zipf's
\cite{zipf45meaning, Zipf1949a} and that have probably
been shadowed by the best-known Zipf's law for word frequencies \cite{Zipf1949a}.
% (tret perque la fras no te sentit) As in any statistical study, he has prevailed the perspective of forest above the tree, so we have sought if there were general correlations  among the studied variables.
Our analysis of the correlation between brevity
(measured in number of characters, phonemes and syllables)
and polysemy (number of synsets) versus word %lemma
frequency
was conducted with three correlation tests with varying assumptions and robustness.
Pearson correlation is a measure of linearity while the Spearman correlation
and Kendall correlation are able to capture monotonic non-linear
dependencies as we have explained in Section \ref{sec:methods}.
Our analysis confirms that a positive correlation between the frequency of
the %lemmas
words and the number of synsets (consistent with the meaning-frequency law \cite{zipf45meaning})
and a negative correlation between the length of the %lemmas
words and their frequency
(consistent with the law of abbreviation \cite{Zipf1949a}) arises under
different definitions of the variables.
In all cases, we find correlations whose sign matches the expected sign.
%\textcolor{green}{FORA: but significance is not achieved in all cases}.
In addition, all correlations are significant
except the Pearson correlations in the meaning-frequency law for Dutch and Spanish.
%From the perspective of the significance of the correlations,
%Tables \ref{corENGchi}, \ref{corENGwik}, \ref{corDUTchi}, \ref{corDUTwik},
%\ref{corSPAchi} and \ref{corSPAwik} show that the tendency
%of length to decrease as frequency increases
%(the law of abbreviation) is a stronger statistical
%law than the tendency of the number of synsets to increase as frequency increases
%(the meaning-frequency law).
%Concerning the meaning-frequency law,
%we find significant positive correlations for the two measures of frequency in all cases,
%\textcolor{red}{with some marginal exceptions in the Pearson correlations in
%Dutch and Spanish}.
This behaviour could be due to (a) the lower capacity of the Pearson correlation
to detect non-linear dependencies compared to Spearman and Kendall correlations or,
(b) the fact that English exhibits a larger sample size
than those two languages (Table \ref{tab:analyzed-words}).

In optimization models of the law of abbreviation,
length is regarded as a proxy for the energetic cost of
the word \cite{Ferrer2012d,Ferrer2015a}.
Then one expects that a better measure of
energetic cost would give a stronger correlation with frequency.
Our meta-analysis of the correlation
between frequency and length has shown
that this correlation is slightly stronger when length
is measured with characters than in phonemes
in most cases in English,
and that characters and phonemes are stronger that
syllables in both English and Dutch.
In Spanish, no clear pattern arises,
which is consistent with the classical
view of Spanish as a more transparent language
than English \cite{nash1977comparing} or Dutch \cite{leufkens2015}.
Thus, the grapheme to phoneme conversion is easier in Spanish
than in English or Dutch \cite{leufkens2015}.
The degree of transparency of a language is defined as the extent to which a language
maintains one-to-one relations between units from different dimensions, e.g., phonemes versus graphemes.
Transparency is tied to the notion of ``\textit{simplicity}" in accounting
for acquisition data (see \cite{leufkens2015} for a review).
Transparency facilitates reading. Then, learning to read in a
transparent orthography imposes fewer constraints
than learning to read in a more opaque writing system \cite{Ijalba2015FirstLG}.
%Consequently, there are minor reading problems in transparent languages \cite{Ijalba2015FirstLG}.

%When involving the three measures of length (number of characters,
%number of phonemes and number of syllables),
%Our meta-analysis of the correlation between frequency and length shows that phonemic length usually yields stronger negative correlations, specially in English. % \textcolor{red}{(Jaume check??? RESPOSTA:when we use wikipedia frequency, pearson correlation reveals that the number of characters is more correlated than the number of syllables, but just with a very small difference (less than 0.01)}.
The fact that the correlation between frequency and number of syllables
tends to be weaker than correlations with other measures of length
does not imply that syllables are a worse measure of length or energetic cost.
It could be simply due to the fact that ties of length values are
easier to obtain with syllabic length, a fact that is expected to yield weaker
correlations and higher p-values as we have shown in Section \ref{ties_subsection}.
% But in order to assess the magnitude of this effect, only p-values and the different coefficients determined may not be sufficient evidence \cite{Baayen2001}, so we do a nonparametric regression function  (local polynomial regression) to the frequency-polysemy and to the frequency-length relationships.

Interestingly, we have not found any remarkable qualitative difference
in the analysis of correlations for adults versus children
in the CHILDES database, suggesting that both child speech and the infant-directed speech or
child-directed speech (the so-called \textit{motherese}) \cite{motheresereview2013} seem
to show the same general statistical biases in the use of more frequent words
(that tend to be shorter and more polysemous),
confirming the results of our previous test in \cite{pilsen}
where adults were split into three different roles,
mother, father and investigator, instead of being considered
together in a single class as in this present paper.
%\textcolor{red}{Jaume: revisa la seccio resultats per veure si aixo que es diu te sentit / val la pena esmentar-ho}
%With this regard, our results agree with Zipf's pioneering discoveries,
%independently from the corpora analyzed and independently
%from the source used to measure the linguistic variables.

Our analyses have shown the robustness of these Zipfian patterns
	from the standpoint of a correlation analysis. Such robustness provides
	support to Zipf's hypothesis that these laws originate from abstract principles,
	e.g., functional pressures (\emph{least effort} as he would put it), that are
	consistent with modern formalizations as a compression principle for the law
	of abbreviation \cite{Ferrer2012d,Ferrer2015a} or a biased random walk over the
	mapping words into meanings for the origins of Zipf's meaning frequency law
	\cite{Ferrer2017b}. This theoretical approaches strongly suggest that it might
	be possible to provide a coherent and parsimonious explanation for the laws
	 we have examined in this article and other laws such as Zipf's law for word
	 frequencies \cite{Ferrer2016compression} or Menzerath's law \cite{Gustison2016a}.
The need for an abstract standpoint is not only suggested by our analyses but also by patterning consistent with these laws in human language in different conditions, e.g., sign language \cite{Boerstell2016a}, Kanji or Chinese characters
\cite{Sanada2008a,Wang2015a}, and also in animal communication
\cite{Ferrer2012d,Ferrer2009f,Hobaiter2014a}. % pendent: canviar Ferrer2012d per article de la Raphaela un cop estigui disponible ??? }.

Our work offers many possibilities for future research.

First, expanding the set of languages to include languages from other families
(i.e. not Indo-European languages)
and the set of lexical databases employed (e.g., \cite{Duchon2013a}). 
As for the latter, the challenge is to find sources that allow to deal 
with different languages homogeneously.
% (Ramon) M'ho he carregat per que es molt vague i el poc que s'enten suposa disparar-nos un tret a la cama (es com dir que no ho hem fet gens be i sabem perque)...
% Second, the use of more fine-grained statistical techniques
% that allow: (1) to unveil differences between sources or
% between kinds of speakers,
% (2) to verify if the statistical tendencies shown here are general or not,
% and (3) to explain the variations that are displayed in which our
% hypotheses hold.

Second, considering different
definitions of the same variables. For instance,
a limitation of our study is the fact that we define word length using discrete units: number of syllables, number of phonemes or number of characters. Future research could benefit from viewing length as a continuous variable, e.g. the (average) duration in time of the word, because that may yield a better estimate of the actual energetic cost of a word and also because our research on language laws is to some extend limited by the information that is transcribed and the writing conventions, that add some degree of arbitrariness. These limitations have been overcome to a large extent in novel investigations of language laws in pure voice \cite{GonzalezTorre2016}.

% ho trec pero potser s'atreveix a adaptar-ho a la nova versio (Ramon)
% An accurate measurement of brevity would require detailed acoustical information
% that is missing in raw written transcripts\cite{GonzalezTorre2016}
% or using more sophisticated methods of computation, for instance,
% to calculate number of phonemes and syllables according to \cite{Altmann2015a}.
% However, the relationship between the duration of phonemes and graphemes
% is well-known and in general longer words have longer durations:
% grapheme-to-phoneme conversion is still a hot topic of research,
% due to the ambiguity of graphemes with respect to their pronunciation
% that today supposes a difficulty in speech technologies \cite{Razavi_SPECOM_2016}.
% In order to improve the frequency measure,
% we would consider the use of alternative databases, e.g., the frequency of English words in Wikipedia \cite{GREFENSTETTE16.624}.

Third, our work can be extended including other linguistic variables such as homophony, i.e. words with different origin
(and \textit{a priori} different meaning) that have converged
to the same phonological form. This extension would require to trace the history of each word, under a dynamical and lexical perspective, following the connection between brevity of words and homophony that Jespersen (1933)
suggested in his seminal work \cite{Jespersen1933} and that has been confirmed more recently \cite{Ke2006,FenkOczlonFenk2010} as a strong association between shortness of words, token frequency and homophony \cite{FenkOczlonFenk2010}. If polysemy is taken to be a form of motivated homophony, by which a word has two or more related meanings, but with probably different representation than homophones (for which different meanings are \textit{"stored separately"} \cite{plosone2016homophony}) in a semantic space, both phenomena will only be distinguishable if we analyze and segment directly voice signals or, as said before, we do that under a diachronic approach.  In any case, we must be aware of the limitations of synchronic approaches when homophony or homography are studied, the later being indistinguishable from polysemy in the present article.

Fourth, a parametric study of these laws with the help of power-law like functions \cite{Naranan1998,Altmann1980a}. In Section \ref{law_of_meaning_distribution_section}, we have shown some challenges of that kind of investigation. We do think that such investigation is very needed and worthwhile. We have just argued it is not as simple as commonly believed.

% \textcolor{red}{
% Finally, our methodology should be applied and extended to other magnitudes such as contextual diversity,
% which could be more relevant than the mere word frequency in some lexical tasks (\cite{Vergara-Martínez2017,adelman2006}),
% thus expanding the Zipfian perspective in psycholinguistics.
% }

Finally, future work should bridge the gap between our classic Zipfian 
perspective and psycholinguistics. We suggest a couple of ways. 
First, an exploration of the structural differences between common 
and rare words \cite{LANDAUER1973119}. 
Second, an application of our methodology to other magnitudes 
such as contextual diversity, which may be more relevant than 
the mere word frequency in some lexical tasks \cite{Vergara-Martínez2017,adelman2006}.

 \medskip

\begin{small}
\section*{Acknowledgments}
The authors thank Pedro Delicado for his helpful comments.
This research work was supported by the grant SGR2014-890 (MACDA) 
and the recognition 2017SGR-856 (MACDA) from AGAUR (Generalitat de Catalunya), 
and also the grants TIN2014-57226-P (APCOM), TIN2017-89244-R (MACDA) 
and TIN2016-77820-C3-3-R (GRAPH-MED) from 
MINECO (Ministerio de Economia, Industria y Competitividad).
\end{small}

\appendix
\section{Figures}
\label{sec:grafiques}

Figures \ref{fig:freqchildes_vs_nsynwordnet}, 
\ref{fig:freqwiki_vs_nsynwordnet}, 
\ref{fig:freqchildes_vs_nlletres}, 
\ref{fig:freqwiki_vs_nlletres}, 
\ref{fig:freqchildes_vs_numphon},
\ref{fig:freqwiki_vs_numphon}, 
\ref{fig:freqchildes_vs_numsylab} and 
\ref{fig:freqwiki_vs_numsylab} 
show the results obtained in this article for English, Dutch and Spanish.

\ifarxiv
\begin{figure}[htp!]
	\centering
	\includegraphics[width=\textwidth]{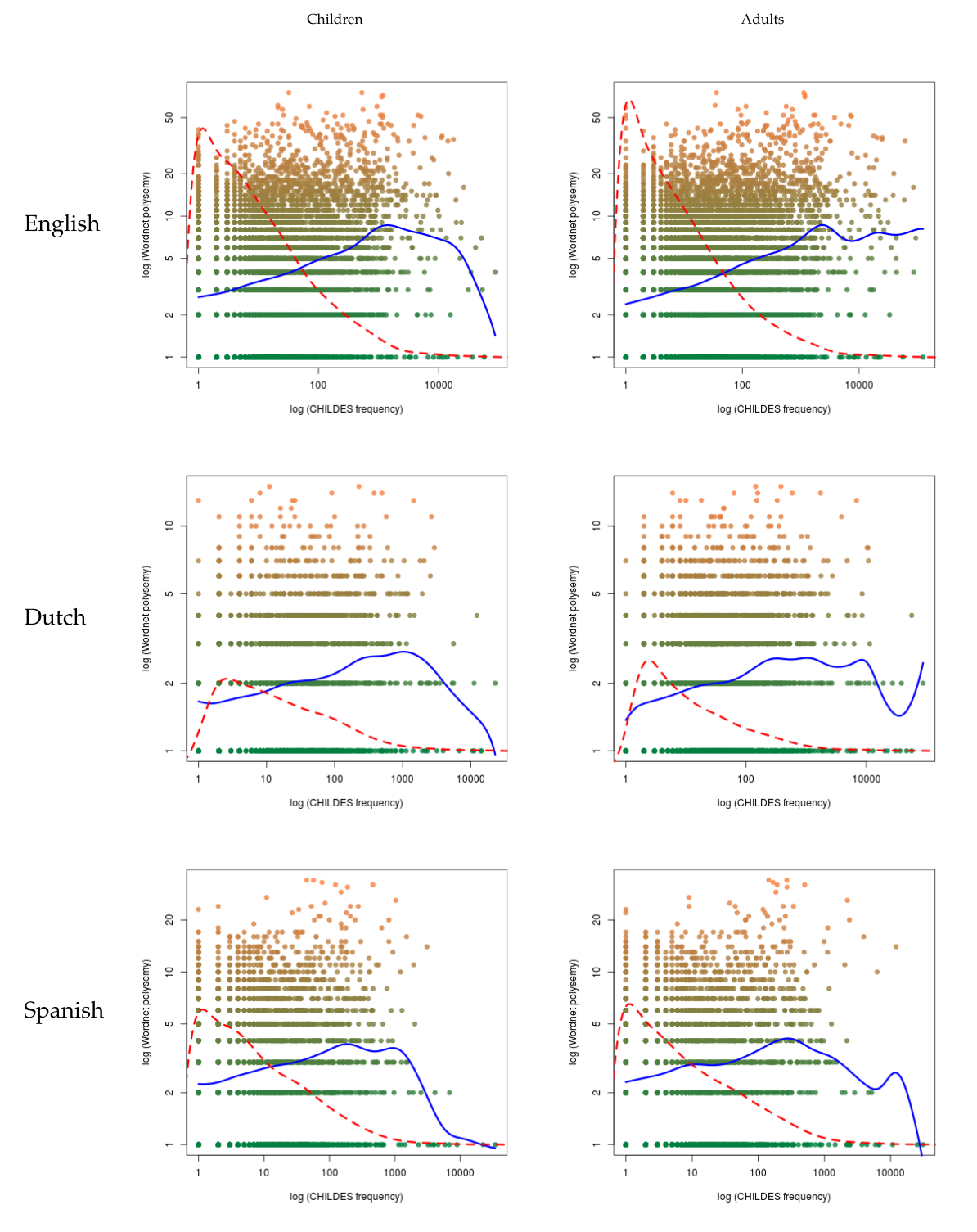}
	\caption{WordNet polysemy versus CHILDES frequency 
		in double logarithmic scale. 
		The color indicates the density of points: dark green is the
		highest possible density. 
		A smoothed curve (blue line) and a curve proportional 
		to the probability density of values of the $x$-axis (red dashed line) is also shown.}
	\label{fig:freqchildes_vs_nsynwordnet}
\end{figure}

\begin{figure}[htp!]
	\centering
	\includegraphics[width=\textwidth]{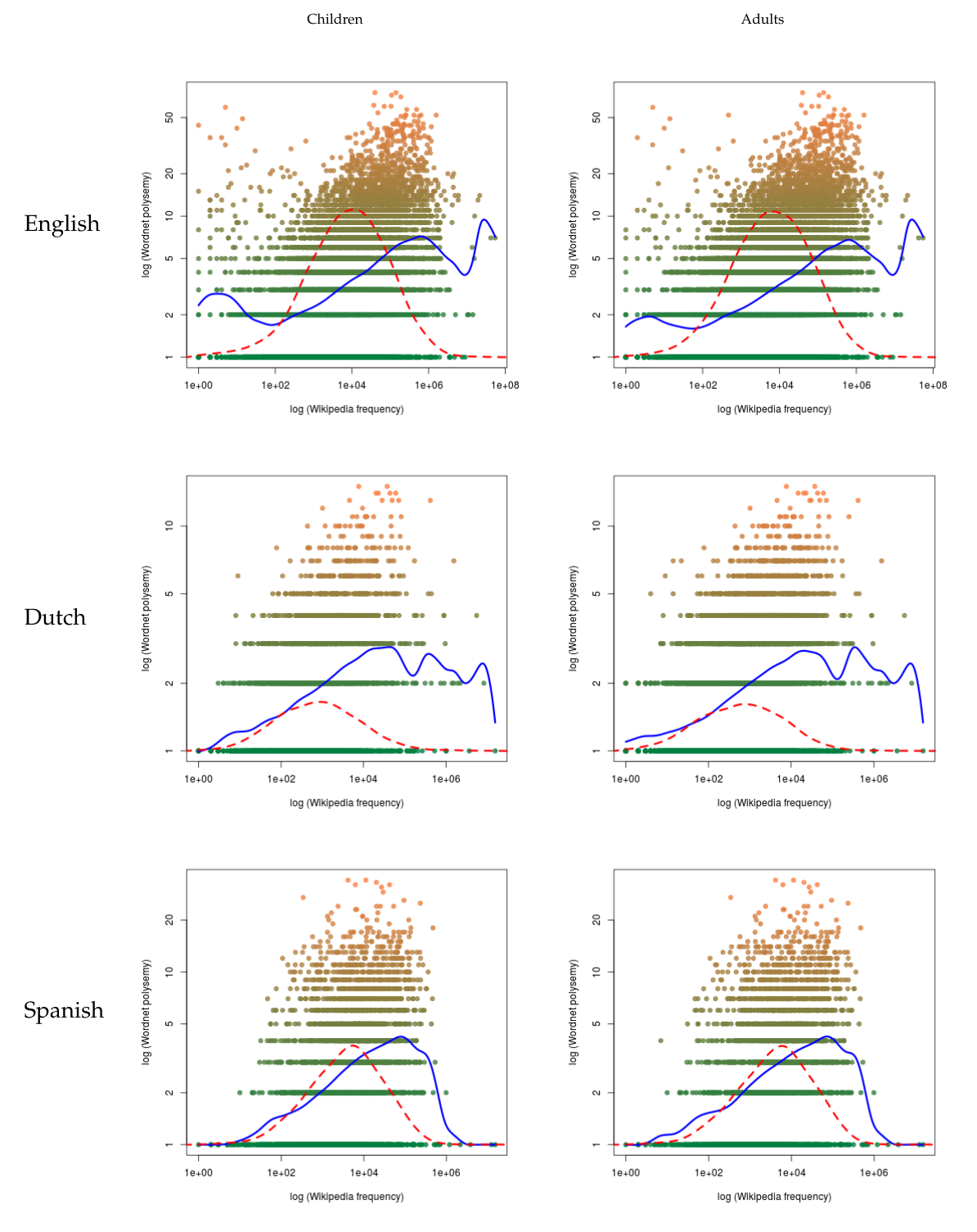}
	\caption{WordNet polysemy versus Wikipedia frequency. 
		The format is the same as in Fig. \ref{fig:freqchildes_vs_nsynwordnet}. }
	\label{fig:freqwiki_vs_nsynwordnet}
\end{figure}

\begin{figure}[htp!]
	\centering
	\includegraphics[width=\textwidth]{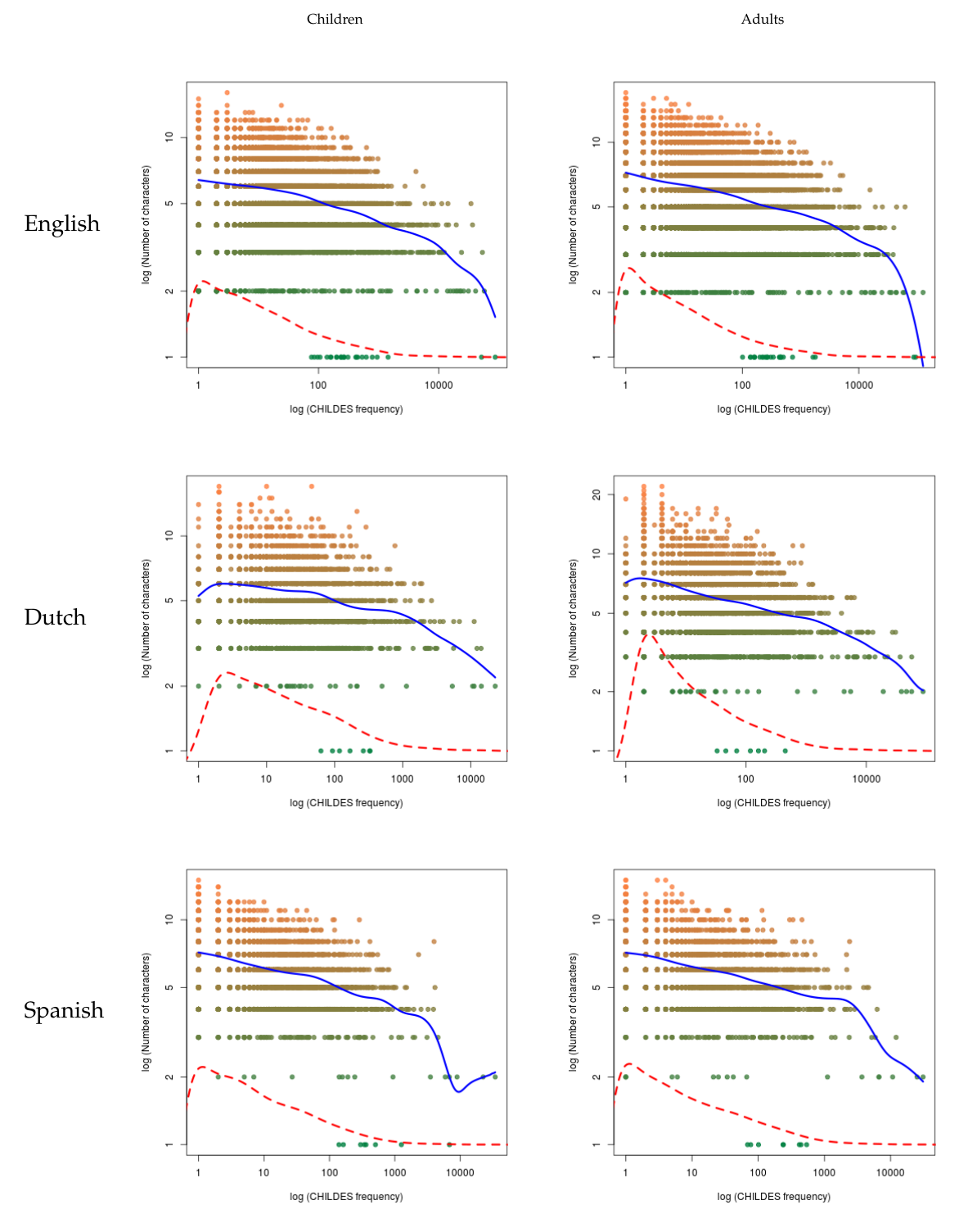}
	\caption{Number of characters versus CHILDES frequency. 
		The format is the same as in Fig. \ref{fig:freqchildes_vs_nsynwordnet}. }
	\label{fig:freqchildes_vs_nlletres}
\end{figure}

\begin{figure}[htp!]
	\centering
	\includegraphics[width=\textwidth]{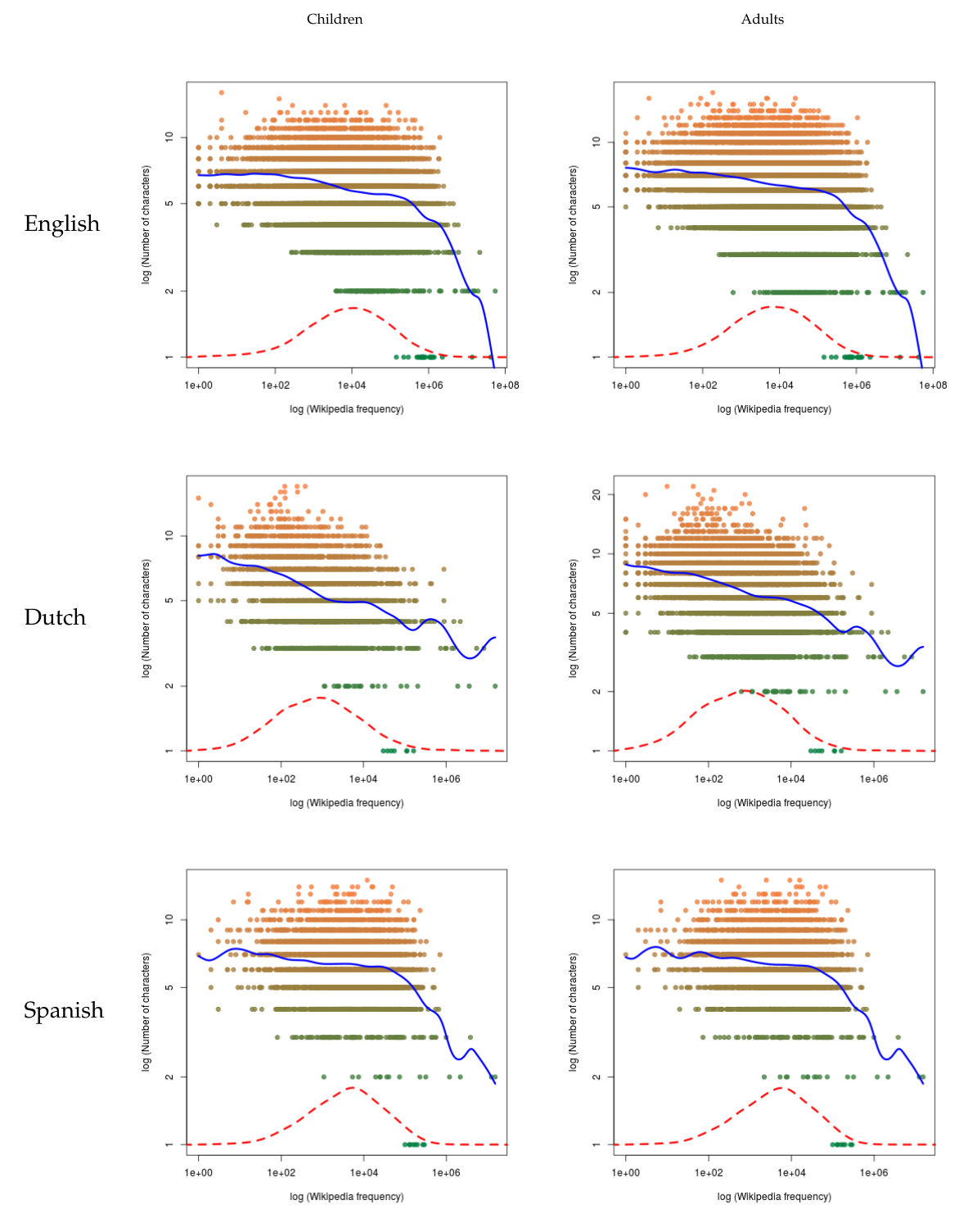}
	\caption{Number of characters versus Wikipedia frequency. 
		The format is the same as in Fig. \ref{fig:freqchildes_vs_nsynwordnet}. }
	\label{fig:freqwiki_vs_nlletres}
\end{figure}

\begin{figure}[htp!]
	\centering
	\includegraphics[width=\textwidth]{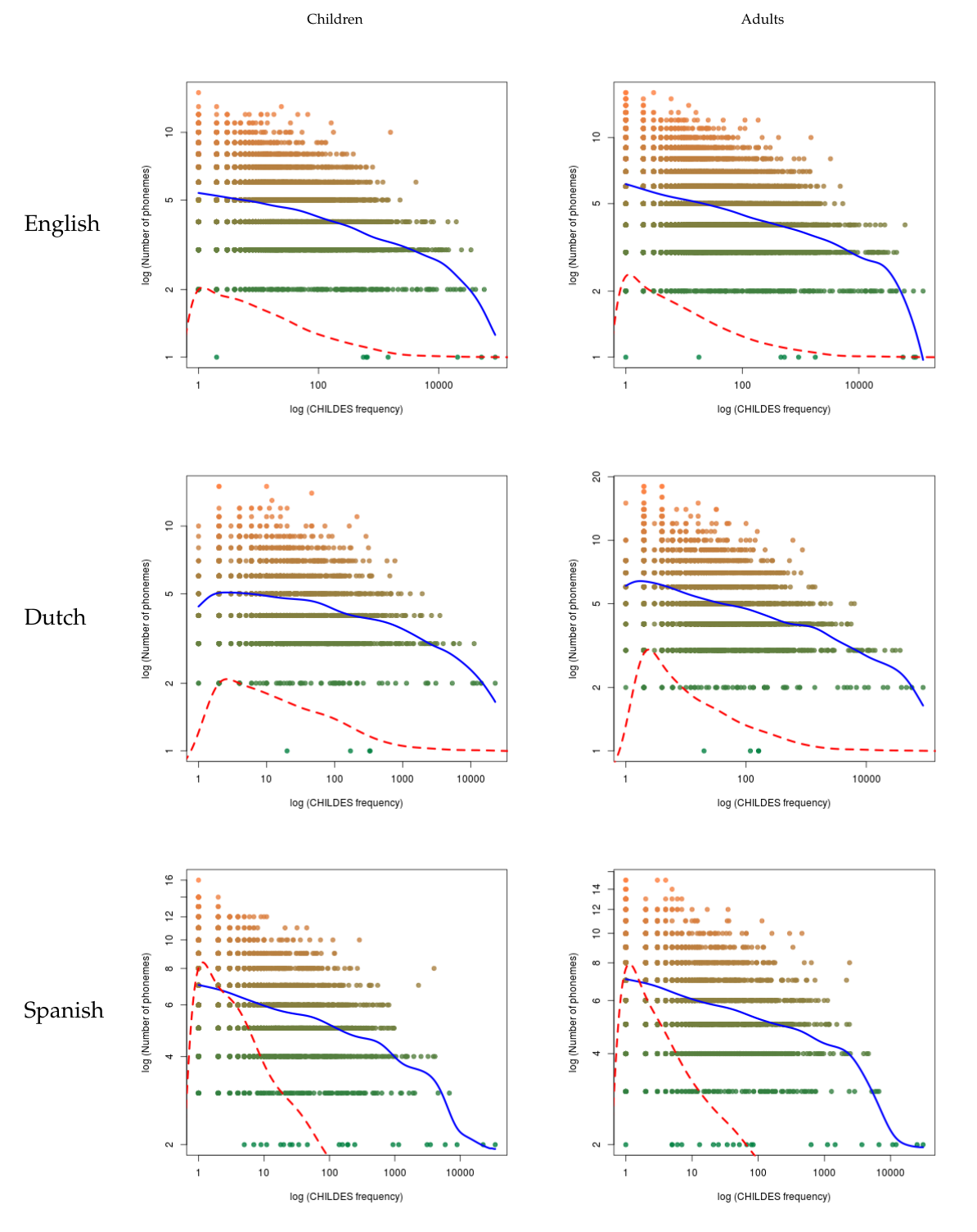}
	\caption{Number of phonemes versus CHILDES frequency. 
		The format is the same as in Fig. \ref{fig:freqchildes_vs_nsynwordnet}. }
	\label{fig:freqchildes_vs_numphon}
\end{figure}

\begin{figure}[htp!]
	\centering
	\includegraphics[width=\textwidth]{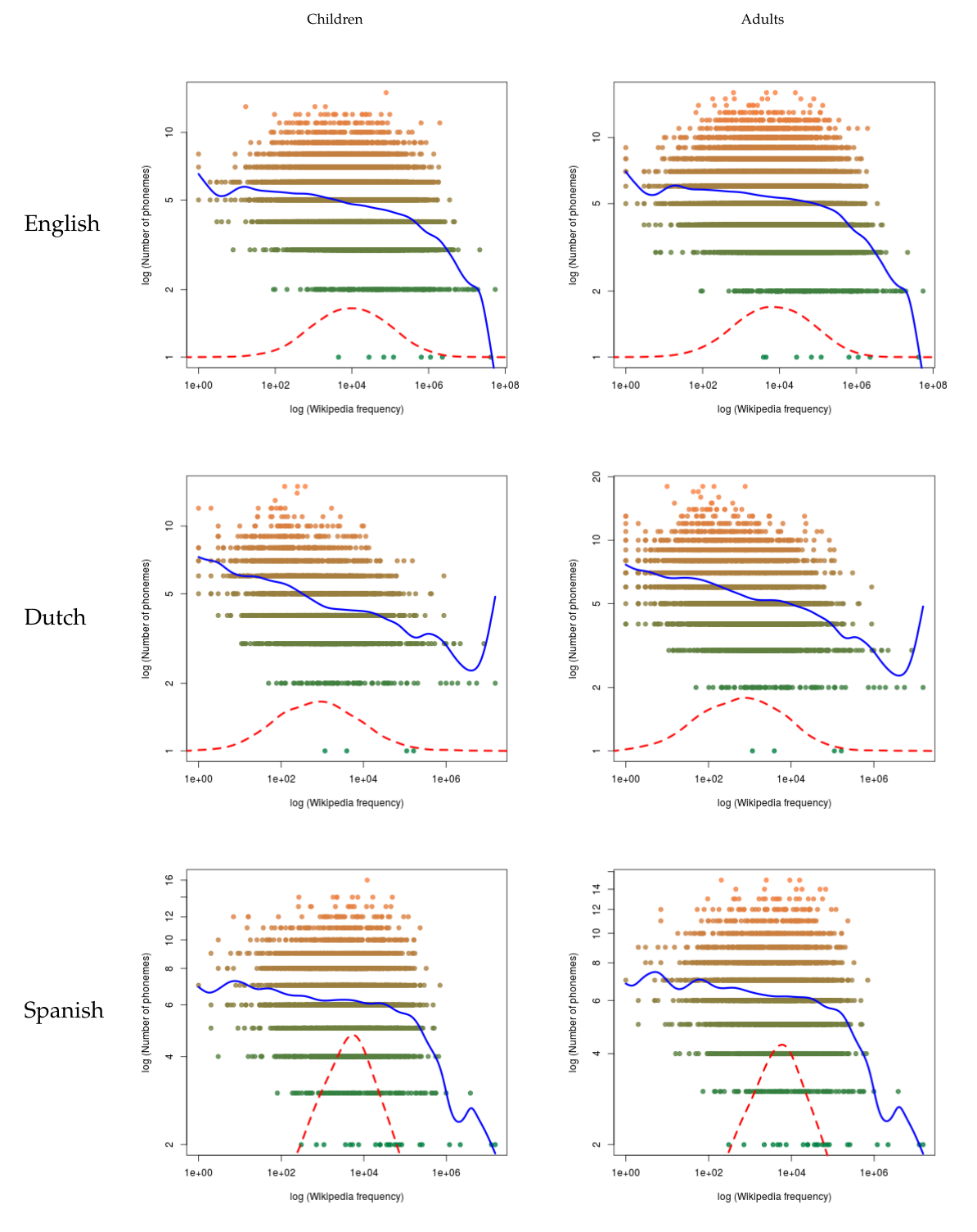}
	\caption{Number of phonemes versus Wikipedia frequency. 
		The format is the same as in Fig. \ref{fig:freqchildes_vs_nsynwordnet}. }
	\label{fig:freqwiki_vs_numphon}
\end{figure}

\begin{figure}[htp!]
	\centering
	\includegraphics[width=\textwidth]{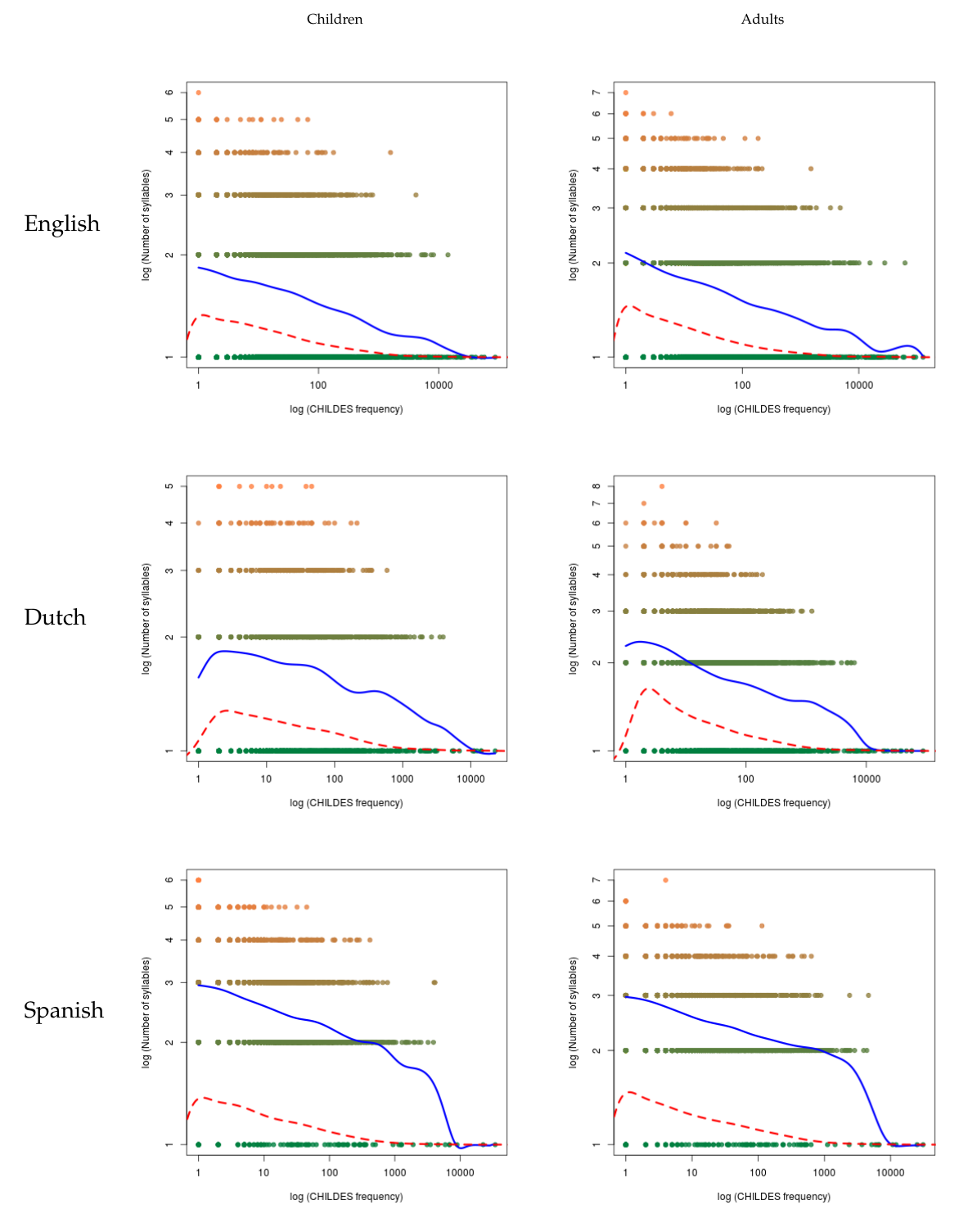}
	\caption{Number of syllables versus CHILDES frequency. 
		The format is the same as in Fig. \ref{fig:freqchildes_vs_nsynwordnet}. }
	\label{fig:freqchildes_vs_numsylab}
\end{figure}

\begin{figure}[htp!]
	\centering
	\includegraphics[width=\textwidth]{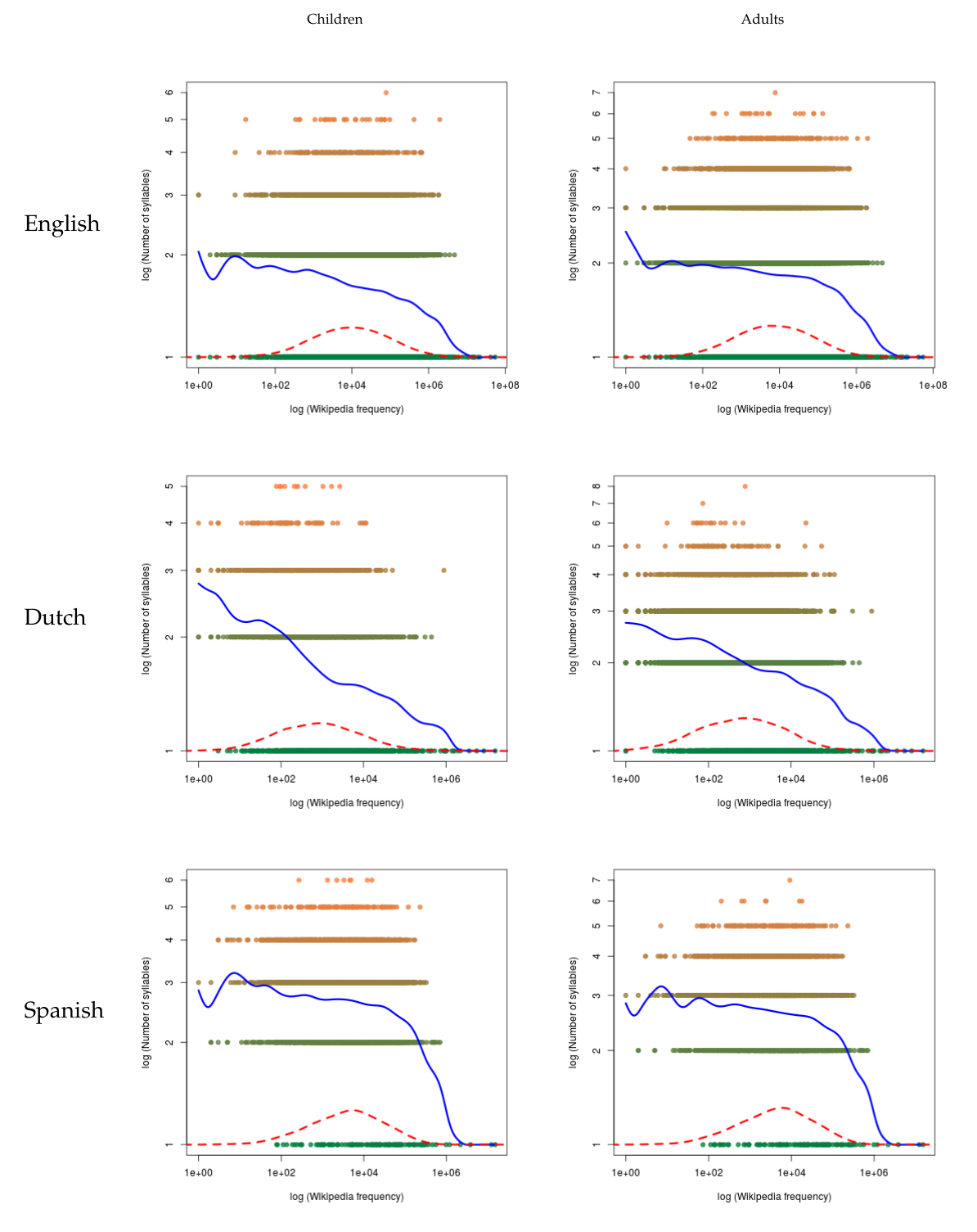}
	\caption{Number of syllables versus Wikipedia frequency. 
		The format is the same as in Fig. \ref{fig:freqchildes_vs_nsynwordnet}. }
	\label{fig:freqwiki_vs_numsylab}
\end{figure}

\fi

In all these plots, frequency is placed on the x-axis for at least three reasons. First, frequency is given (frequency is assumed to be constant) while length is variable In information theory (coding theory in particular) \cite{Cover2006a}. In the problem of compression, one aims to minimize the average length of codes given probabilities (estimated as relative frequencies). Information theory predicts the length of a code as a function of its frequency \cite[p. 111]{Cover2006a} or its frequency rank \cite{Ferrer2015a}. Therefore, when plotting length versus frequency, it makes sense to put it on the x-axis. Second, word frequency is a fundamental variable in psycholinguistics to predict language processing costs \cite{Brown1966a,Connine1990a}.
The third reason comes from the popular Zipf's law for word frequencies. Although Zipf's law is usually plotted as frequency as function of rank (following Eq. \ref{Zipfs_law_equation}), a sister (but not identical) plot, 
the so called frequency spectrum, consists of showing the number of distinct words as a function of frequency \cite{Tuldava1996}. The second plot (frequency on the x-axis) is preferred by various authors for investigating the distribution of word frequencies (e.g., \cite[p. 298]{Naranan1992c}, \cite[p. 3]{Moreno2016a}).

\section{Information about CHILDES}
\label{sec:appendixchildes}

Tables \ref{tab:childes_british}, \ref{tab:childes_american}, \ref{tab:childes_dutch} and \ref{tab:childes_spanish} show the CHILDES corpora used in this article for English, Dutch and Spanish.

%\textcolor{red}{Crec que seria millor traslladar les taules 2,3,4 i 5 a un appendix, per facilitar la lectura de l'article}

\begin{table}[H]
	\begin{center}
		\resizebox{\textwidth}{!}{
		\begin{tabular}{|c|c|c|p{4cm}|}
			\hline
			Corpus & Age Range & \# children & Comments \\
			\hline
			\hline
			Lara \cite{Rowland2006a} & 1;9 -- 3;3 & 1 & Longitudinal case study \\
			\hline
			Manchester \cite{Theakston2001a} & 1;8 -- 3;0 & 12 & 12 English children recorded weekly for the period of a year \\
			\hline
			Wells \cite{Wells1981a} & 1;6 -- 5;0 & 32 & Large study of the language of British preschool children collected at random intervals \\
			\hline
		\end{tabular}}
		\caption{CHILDES database for British English \url{http://childes.psy.cmu.edu/access/Eng-UK/}}
		\label{tab:childes_british}
	\end{center}
\end{table}

\begin{table}[H]
	\begin{center}
		\resizebox{\textwidth}{!}{			
		\begin{tabular}{|p{2.5cm}|c|c|p{4cm}|}
			\hline
			Corpus & Age Range & \# children & Comments \\
			\hline
			\hline
			Bloom 1970 \cite{Bloom1974a, Bloom1975a, Bloom1970a}
			& 1;9 -- 3;2 & 2 & A large longitudinal study of one child
			with a few samples for another. Gia was excluded because age information is not reported for her. \\
			\hline
			\multirow{3}{*}{Brown \cite{Brown1973a}} & Adam 2;3 -- 4;10 & \multirow{3}{*}{3} & Large longitudinal study \\
			& Eve 1;6 -- 2;3 & & of three children: Adam \\
			& Sarah 2;3 -- 5;1 & & 55 files, Eve 20 and Sarah 139 \\
			\hline
			Kuczaj \cite{Kuczaj1977a} & 2;4 -- 5;0 & 1 & Diary study in the home environment \\
			\hline
			MacWhinney \cite{Childes_Database_Manual_for_American_English}
			& Ross 2;6 -- 8;0 & \multirow{2}{*}{2} & Diary study of the development of two
			brothers recorded in spontaneous situations \\
			& Mark 0;7 -- 5;6 & & \\
			\hline
			Providence \cite{Demuth2006a} & 1 -- 3 & 5 & Ethan was excluded
			because he was diagnosed with Asperger's Syndrome at the
			age of 5. \\
			\hline
			Sachs \cite{Sachs1983a} & 1;1 -- 5;1 & 1 & Longitudinal naturalistic study \\
			\hline
			Suppes \cite{Suppes1974a} & 1;11 -- 3;11 & 1 & Longitudinal study of a single child \\
			\hline
		\end{tabular}}
		\caption{CHILDES database for American English \url{http://childes.psy.cmu.edu/access/Eng-NA/}}
		\label{tab:childes_american}
	\end{center}
\end{table}

\begin{table}[H]
	\begin{center}
		\resizebox{\textwidth}{!}{
			\begin{tabular}{|p{2.5cm}|c|c|p{4cm}|}
				\hline
				Corpus & Age Range & \# children & Comments \\
				\hline
				\hline
				BolKuiken \cite{doi:10.3109/02699209008985472} & 1;7 -- 3;7 & 47 & Dutch normal controls \\
				\hline
				CLPF \cite{Fikkert1994, Levelt1994} & 1;0 -- 2;11 & 12 & PHONBANK, longitudinal study with 20,000 utterances \\
				\hline
				Groningen \cite{Bol1995a} & 1;5 -- 3;7 & 6 & 'Iris'
				was removed because she subsequently displayed
				delay in language development due to hearing problems. 'Iri'
				(ending with no 's') was also excluded (this person was very
				likely a misspelling of 'Iris' because he/she was in the same
				subdirectory of 'Iris' and was the only target child in the only
				file where it appeared). \\
				\hline
				Schaerlaekens \cite{Schaerlaekens1973a} & 1;8 -- 2;10 ,
				1;10 -- 3;1 & 6 & \\
				\hline
				\multirow{2}{*}{van Kampen \cite{VanKampen1994a}}
				& Laura 1;9 -- 5;10 & \multirow{2}{*}{2} & \\
				& Sarah 1;6.16 -- 6;0 & & \\
				\hline
		\end{tabular}}
		\caption{CHILDES database for Dutch \url{http://childes.psy.cmu.edu/access/Dutch/}}
		\label{tab:childes_dutch}
	\end{center}
\end{table}

\begin{table}[H]
	\begin{center}
		\resizebox{\textwidth}{!}{
			\begin{tabular}{|p{2.5cm}|c|c|p{4cm}|}
				\hline
				Corpus & Age Range & \# children & Comments \\
				\hline
				Aguirre \cite{Aguirre2000} & 1;7-2;10 & 1 & \\
				\hline
				BecaCESNo & 3;6-11;6 & 40 & \\
				\hline
				ColMex & 6;0-7;0 & 30 & Mexican Spanish, picture and procedural description \\
				\hline
				DiezItza \cite{DiezItza1995} & 3;0-3;11 & 20 & \\
				\hline
				FernAguado & 3;0-4;0 & 50 & \\
				\hline
				Hess \cite{Hess2003} & 6;0-12;0 & 24 & \\
				\hline
				JacksonThal \cite{Jackson1993, Thal1993} & 0;10-3;0 & 202 & Cross-sectional data from Queretaro, San Diego, and Santa Barbara\\
				\hline
%				Koine & 1;6-4;5 & 71 & \\
%				\hline
				Linaza \cite{Linaza1981} & 2;0-4;0 & 1 & \\
				\hline
				LlinasOjea & 0;11-3;02 & 1 & Longitudinal study of two children in Asturias, but only Yasmin is considered. \\
				\hline
				Marrero & 1;8-8;0 & 3 & Longitudinal study of Spanish children from the Canaries\\
				\hline
				Nieva & 1;8-2;3 & 1 & \\
				\hline
				Ornat \cite{Ornat1994} & 1;7-4;0 & 1 & \\
				\hline
				Remedi & 1;11-2;10 & 1 & \\
				\hline
				Romero \cite{Romero1992} & 2;0 & 1 & Mexican Spanish \\
				\hline
				SerraSole & 1;4-3;10 & 1 & \\
				\hline
				Shiro \cite{Shiro1997} & 6;0-9;0 & 113 & Narratives from Venezuelan children \\
				\hline
%				Vila & 0;11-4;08 & 1 & \\
%				\hline
		\end{tabular}}
		\caption{CHILDES database for Spanish \url{http://childes.psy.cmu.edu/access/Spanish/}}
		\label{tab:childes_spanish}
	\end{center}
\end{table}

\section{Correlations}
\label{sec:appendixcorrelations}

Tables \ref{corENGchi}, \ref{corENGwik}, \ref{corDUTchi}, \ref{corDUTwik}, \ref{corSPAchi} and \ref{corSPAwik} show the correlations between Frequency and Polysemy, and Frequency and Length measures for English, Dutch and Spanish.

\corrEngCHILDES		% corENGchi
\corrEngWiki		% corENGwik

\corrNldCHILDES		% corDUTchi
\corrNldWiki		% corDUTwik

\corrSpaCHILDES		% corSPAchi
\corrSpaWiki		% corSPAchi

\section{Tables of Steiger's tests}
\label{sec:appendixsteiger}

Tables \ref{nlletresnphon}, \ref{nlletresnsyllab} and \ref{nphonnsyllab}
show the results of Steiger's test 
between number of characters and number of phonemes, 
number of characters and number of syllables, 
and number of phonemes and number of syllables for English, Dutch and Spanish.

\lletresphon 		% nlletresnphon
\lletressyllab 		% nlletresnsyllab
\phonsyllab 		% nphonnsyllab

%\bibliographystyle{plain}
%\bibliographystyle{splncs03}
%\bibliography{17-csl}

\newpage
\section*{References}
\bibliographystyle{elsarticle-num}
\bibliography{csl17}

\begin{thebibliography}{10}
\expandafter\ifx\csname url\endcsname\relax
  \def\url#1{\texttt{#1}}\fi
\expandafter\ifx\csname urlprefix\endcsname\relax\def\urlprefix{URL }\fi
\expandafter\ifx\csname href\endcsname\relax
  \def\href#1#2{#2} \def\path#1{#1}\fi

\bibitem{Zipf1949a}
G.~K. Zipf, Human behaviour and the principle of least effort, Addison-Wesley,
  Cambridge (MA), USA, 1949.

\bibitem{Zipf1968a}
G.~K. Zipf, The Psycho-Biology of Language: an Introduction to Dynamic
  Psychology, MIT Press, Cambridge, MA, USA, 1968, originally published in 1935
  by Houghton Mifflin - Boston - MA - USA.

\bibitem{zipf45meaning}
G.~K. Zipf, {The Meaning-Frequency Relationship of Words}, Journal of General
  Psychology 1945~(33) (1945) 251--256.

\bibitem{Naranan1998}
S.~Naranan, V.~K. Balasubrahmanyan, Models for power law relations in
  linguistics and information science, J. Quantitative Linguistics 5~(1-2)
  (1998) 35--61.

\bibitem{DBLP:dblp_journals/corr/Ferrer-i-Cancho14e}
R.~{Ferrer-i-Cancho}, \href{http://arxiv.org/abs/1412.2486}{Optimization models
  of natural communication}, in: Journal of Quantitative Linguistics, Vol.~25,
  2014.
\newline\urlprefix\url{http://arxiv.org/abs/1412.2486}

\bibitem{1367-2630-15-9-093033}
F.~Font-Clos, G.~Boleda, A.~Corral,
  \href{http://stacks.iop.org/1367-2630/15/i=9/a=093033}{A scaling law beyond
  {Zipf's} law and its relation to {Heaps'} law}, New Journal of Physics 15~(9)
  (2013) 093033.
\newline\urlprefix\url{http://stacks.iop.org/1367-2630/15/i=9/a=093033}

\bibitem{10.1371/journal.pone.0129031}
A.~Corral, G.~Boleda, R.~{Ferrer-i-Cancho}, Zipf's law for word frequencies:
  Word forms versus lemmas in long texts, PLoS ONE 10~(7) (2015) 1--23.
\newblock \href {http://dx.doi.org/10.1371/journal.pone.0129031}
  {\path{doi:10.1371/journal.pone.0129031}}.

\bibitem{Ferrer2014d}
R.~{Ferrer-i-Cancho}, The meaning-frequency law in {Zipfian} optimization
  models of communication, Glottometrics 35 (2016) 28--37.

\bibitem{grzybek2006contributions}
P.~Grzybek, Contributions to the science of text and language: word length
  studies and related issues, Vol.~31, Springer Science \& Business Media,
  2006.

\bibitem{Strauss2007a}
U.~Strauss, P.~Grzybek, G.~Altmann, Word length and word frequency, Springer,
  Dordrecht, 2007, pp. 277--294.

\bibitem{Ilgen2007a}
B.~Ilgen, B.~Karaoglan, Investigation of {Zipf}'s ``law-of-meaning'' on
  {Turkish} corpora, in: 22nd International Symposium on Computer and
  Information Sciences (ISCIS 2007), 2007, pp. 1--6.

\bibitem{Ferrer2012d}
R.~{Ferrer-i-Cancho}, A.~Hern\'{a}ndez-Fern\'{a}ndez, D.~Lusseau,
  G.~Agoramoorthy, M.~J. Hsu, S.~Semple, Compression as a universal principle
  of animal behavior, Cognitive Science 37~(8) (2013) 1565--1578.

\bibitem{Ferrer2017b}
R.~{Ferrer-i-Cancho}, M.~Vitevitch, The origins of {Zipf's} meaning-frequency
  law, Journal of the American Association for Information Science and
  Technology 69 (2018) 1369--1379.

\bibitem{GonzalezTorre2016}
I.~Gonzalez~Torre, B.~Luque, L.~Lacasa, J.~Luque, A.~Hernandez-Fernandez,
  \href{http://www.nature.com/articles/srep43862}{Emergence of linguistic laws
  in human voice}, Scientific reports 7~(43862) (2017) 1--10.
\newblock \href {http://dx.doi.org/10.1038/srep43862}
  {\path{doi:10.1038/srep43862}}.
\newline\urlprefix\url{http://www.nature.com/articles/srep43862}

\bibitem{pilsen}
A.~Hern{\'a}ndez-Fern{\'a}ndez, B.~Casas, R.~{Ferrer-i-Cancho}, J.~Baixeries,
  \href{http://dx.doi.org/10.1007/978-3-319-45925-7_2}{Testing the Robustness
  of Laws of Polysemy and Brevity Versus Frequency}, Springer International
  Publishing, Cham, 2016, pp. 19--29.
\newblock \href {http://dx.doi.org/10.1007/978-3-319-45925-7_2}
  {\path{doi:10.1007/978-3-319-45925-7_2}}.
\newline\urlprefix\url{http://dx.doi.org/10.1007/978-3-319-45925-7_2}

\bibitem{MacWhinney2000a}
B.~MacWhinney, The {CHILDES} project: tools for analyzing talk, 3rd Edition,
  Vol. 2: the database, Lawrence Erlbaum Associates, Mahwah, NJ, 2000.

\bibitem{FellBaum1998a}
C.~Fellbaum, {WordNet: An Electronic Lexical Database}, MIT Press, Cambridge,
  MA, 1998.

\bibitem{GREFENSTETTE16.624}
G.~Grefenstette, {Extracting Weighted Language Lexicons from Wikipedia}, in:
  N.~Calzolari, K.~Choukri, T.~Declerck, S.~Goggi, M.~Grobelnik, B.~Maegaard,
  J.~Mariani, H.~Mazo, A.~Moreno, J.~Odijk, S.~Piperidis (Eds.), Proceedings of
  the Tenth International Conference on Language Resources and Evaluation (LREC
  2016), European Language Resources Association (ELRA), Paris, France, 2016.

\bibitem{motheresereview2013}
C.~Saint-Georges, M.~Chetouani, R.~Cassel, F.~Apicella, A.~Mahdhaoui,
  F.~Muratori, M.-C. Laznik, D.~Cohen,
  \href{http://dx.doi.org/10.1371%2Fjournal.pone.0078103}{Motherese in
  interaction: At the cross-road of emotion and cognition? ({A} systematic
  review)}, PLOS ONE 8~(10).
\newblock \href {http://dx.doi.org/10.1371/journal.pone.0078103}
  {\path{doi:10.1371/journal.pone.0078103}}.
\newline\urlprefix\url{http://dx.doi.org/10.1371%2Fjournal.pone.0078103}

\bibitem{Ide2006}
N.~Ide, Y.~Wilks, \href{http://dx.doi.org/10.1007/978-1-4020-4809-8_3}{Making
  Sense About Sense}, Springer Netherlands, Dordrecht, 2006, pp. 47--73.
\newblock \href {http://dx.doi.org/10.1007/978-1-4020-4809-8_3}
  {\path{doi:10.1007/978-1-4020-4809-8_3}}.
\newline\urlprefix\url{http://dx.doi.org/10.1007/978-1-4020-4809-8_3}

\bibitem{Kilgarriff1992}
A.~Kilgarriff, \href{http://dx.doi.org/10.1007/BF00136981}{Dictionary word
  sense distinctions: An enquiry into their nature}, Computers and the
  Humanities 26~(5) (1992) 365--387.
\newblock \href {http://dx.doi.org/10.1007/BF00136981}
  {\path{doi:10.1007/BF00136981}}.
\newline\urlprefix\url{http://dx.doi.org/10.1007/BF00136981}

\bibitem{Zugarramurdi}
B.~Armstrong, C.~Zugarramurdi, A.~Cabana, J.~Valle~Lisboa, D.~Plaut, Relative
  meaning frequencies for 578 homonyms in two {Spanish} dialects: {A}
  cross-linguistic extension of the {English eDom} norms, Behavior Research
  Methods 48.
\newblock \href {http://dx.doi.org/10.3758/s13428-015-0639-3}
  {\path{doi:10.3758/s13428-015-0639-3}}.

\bibitem{FRAGA20171}
I.~Fraga, I.~Padr\'on, M.~Perea, M.~Comesa{\~n}a,
  \href{http://www.sciencedirect.com/science/article/pii/S0024384116300596}{I
  saw this somewhere else: {The Spanish Ambiguous Words} ({SAW}) database},
  Lingua 185 (2017) 1 -- 10.
\newblock \href
  {http://dx.doi.org/https://doi.org/10.1016/j.lingua.2016.07.002}
  {\path{doi:https://doi.org/10.1016/j.lingua.2016.07.002}}.
\newline\urlprefix\url{http://www.sciencedirect.com/science/article/pii/S0024384116300596}

\bibitem{Altmann1980a}
G.~Altmann, Prolegomena to {Menzerath's} law, Glottometrika 2 (1980) 1--10.

\bibitem{Altmann2015a}
E.~G. Altmann, M.~Gerlach,
  \href{http://dx.doi.org/10.1007/978-3-319-24403-7_2}{Statistical Laws in
  Linguistics}, Springer International Publishing, Cham, 2016, pp. 7--26.
\newblock \href {http://dx.doi.org/10.1007/978-3-319-24403-7_2}
  {\path{doi:10.1007/978-3-319-24403-7_2}}.
\newline\urlprefix\url{http://dx.doi.org/10.1007/978-3-319-24403-7_2}

\bibitem{Font2015a}
F.~Font-Clos, A.~Corral, Log-log convexity of type-token growth in {Zipf}'s
  systems, Phys. Rev. Lett. 114 (2015) 238701.
\newblock \href {http://dx.doi.org/10.1103/PhysRevLett.114.238701}
  {\path{doi:10.1103/PhysRevLett.114.238701}}.

\bibitem{Ferrer2012h}
R.~{Ferrer-i-Cancho}, A.~Hern\'andez-Fern\'andez, J.~Baixeries,
  {\L}.~D\k{e}bowski, J.~Ma\v{c}utek, When is {Menzerath-Altmann} law
  mathematically trivial? {A} new approach, Statistical Applications in
  Genetics and Molecular Biology 13 (2014) 633--644.

\bibitem{Seal1952}
H.~L. Seal, The maximum likelihood fitting of the discrete {Pareto} law,
  Journal of the Institute of Actuaries (1886-1994) 78~(1) (1952) 115--121.

\bibitem{Corral2006a}
A.~Corral, Dependence of earthquake recurrence times and independence of
  magnitudes on seismicity history, Tectonophysics 424 (2006) 177--193.

\bibitem{Bond:Foster:2013}
F.~Bond, R.~Foster, {Linking and Extending an Open Multilingual Wordnet},
  Sofia, 2013.

\bibitem{Postma:Miltenburg:Segers:Schoen:Vossen:2016}
M.~Postma, E.~van Miltenburg, R.~Segers, A.~Schoen, P.~Vossen, Open {Dutch}
  {WordNet}, in: Proceedings of the Eight Global Wordnet Conference, Bucharest,
  Romania, 2016.

\bibitem{Gonzalez-Agirre:Laparra:Rigau:2012}
A.~Gonzalez-Agirre, E.~Laparra, G.~Rigau, Multilingual central repository
  version 3.0: upgrading a very large lexical knowledge base, in: Proceedings
  of the 6th Global WordNet Conference (GWC 2012), Matsue, 2012.

\bibitem{CELEX}
R.~H. Baayen, R.~Piepenbrock, L.~Gulikers, \href{http://celex.mpi.nl}{CELEX}
  (1996).
\newline\urlprefix\url{http://celex.mpi.nl}

\bibitem{Conover1999a}
W.~J. Conover, Practical nonparametric statistics, Wiley, New York, 1999, 3rd
  edition.

\bibitem{Gibbons2010a}
J.~D. Gibbons, S.~Chakraborti, Nonparametric statistical inference, Chapman and
  Hall/CRC, Boca Raton, FL, 2010, 5th edition.

\bibitem{Embrecths2002a}
P.~Embrechts, A.~McNeil, D.~Straumann, Correlation and dependence in risk
  management: properties and pitfalls, in: M.~A.~H. Dempster (Ed.), Risk
  management: value at risk and beyond, Cambridge University Press, Cambridge,
  2002, pp. 176--223.

\bibitem{Steiger1980a}
J.~H. Steiger, Tests for comparing elements of a correlation matrix,
  Psychological Bulletin 87 (1980) 245--251.

\bibitem{Daniels1950a}
H.~E. Daniels, Rank correlation and population models, Journal of the Royal
  Statistical Society, Series B 12 (1950) 171--81.

\bibitem{Durbin1951a}
J.~Durbin, A.~Stuart, Inversions and rank correlations, Journal of the Royal
  Statistical Society, Series B 13 (1951) 303--309.

\bibitem{Ferrer2015a}
R.~{Ferrer-i-Cancho}, C.~Bentz, C.~Seguin,
  \href{http://arxiv.org/abs/1504.04884}{Compression and the origins of
  {Zipf's} law of abbreviation}.
\newline\urlprefix\url{http://arxiv.org/abs/1504.04884}

\bibitem{nash1977comparing}
R.~Nash, \href{https://books.google.es/books?id=ke86OwAACAAJ}{Comparing
  {English} and {Spanish}: Patterns in Phonology and Orthography}, Prentice
  Hall, 1977.
\newline\urlprefix\url{https://books.google.es/books?id=ke86OwAACAAJ}

\bibitem{leufkens2015}
S.~Leufkens, \href{http://hdl.handle.net/11245/1.439561}{Transparency in
  language: a typological study}, LOT, 2015.
\newline\urlprefix\url{http://hdl.handle.net/11245/1.439561}

\bibitem{Ijalba2015FirstLG}
E.~Ijalba, L.~K. Obler, First language grapheme-phoneme transparency effects in
  adult second-language learning, Vol.~27, 2015, pp. 47--70.

\bibitem{Ferrer2016compression}
R.~{Ferrer-i-Cancho}, \href{http://dx.doi.org/10.1002/cplx.21820}{Compression
  and the origins of {Zipf's} law for word frequencies}, Complexity 21 (2016)
  409--411.
\newblock \href {http://dx.doi.org/10.1002/cplx.21820}
  {\path{doi:10.1002/cplx.21820}}.
\newline\urlprefix\url{http://dx.doi.org/10.1002/cplx.21820}

\bibitem{Gustison2016a}
M.~L. Gustison, S.~Semple, R.~{Ferrer-i-Cancho}, T.~Bergman, Gelada vocal
  sequences follow {Menzerath}'s linguistic law, Proceedings of the National
  Academy of Sciences USA 13 (2016) E2750--E2758.
\newblock \href {http://dx.doi.org/10.1073/pnas.1522072113}
  {\path{doi:10.1073/pnas.1522072113}}.

\bibitem{Boerstell2016a}
C.~B\"{o}rstell, T.~H\"{o}rberg, R.~\"{O}stling, Distribution and duration of
  signs and parts of speech in {Swedish Sign Language}, Sign Language \&
  Linguistics 19 (2016) 143--196.

\bibitem{Sanada2008a}
H.~Sanada, Investigations in Japanese historical lexicology, Peust \&
  Gutschmidt Verlag, G\"{o}ttingen, 2008.

\bibitem{Wang2015a}
Y.~Wang, X.~Chen, Structural complexity of simplified {Chinese} characters, in:
  A.~Tuzzi, J.~M. M.~Benesov\'a (Eds.), Recent Contributions to Quantitative
  Linguistics, De Gruyter, 2015, pp. 229--239.

\bibitem{Ferrer2009f}
R.~{Ferrer-i-Cancho}, B.~McCowan, A law of word meaning in dolphin whistle
  types, Entropy 11~(4) (2009) 688--701.
\newblock \href {http://dx.doi.org/10.3390/e11040688}
  {\path{doi:10.3390/e11040688}}.

\bibitem{Hobaiter2014a}
C.~Hobaiter, R.~W. Byrne, The meanings of chimpanzee gestures, Current Biology
  24 (2014) 1596--1600.

\bibitem{Duchon2013a}
A.~Duchon, M.~Perea, N.~Sebasti{\'a}n-Gall{\'e}s, A.~Mart{\'i}, M.~Carreiras,
  {EsPal}: {One-stop} shopping for {Spanish} word properties, Behavior Research
  Methods 45~(4) (2013) 1246--1258.

\bibitem{Jespersen1933}
O.~Jespersen, Monosyllabism in {English}, in: Linguistica: Selected Writings of
  Otto Jespersen, George Allen and Unwin LTD, London, UK, 2007, pp. 574--598.

\bibitem{Ke2006}
J.~Ke, A cross-linguistic quantitative study of homophony, Journal of
  Quantitative Linguistics (2006) 129--159.

\bibitem{FenkOczlonFenk2010}
G.~Fenk-Oczlon, A.~Fenk, Frequency effects on the emergence of polysemy and
  homophony, International Journal Information Technologies and Knowledge 4~(2)
  (2010) 103--109.

\bibitem{plosone2016homophony}
I.~Dautriche, E.~Chemla,
  \href{http://dx.doi.org/10.1371%2Fjournal.pone.0162176}{What homophones say
  about words}, PLOS ONE 11~(9) (2016) 1--19.
\newblock \href {http://dx.doi.org/10.1371/journal.pone.0162176}
  {\path{doi:10.1371/journal.pone.0162176}}.
\newline\urlprefix\url{http://dx.doi.org/10.1371%2Fjournal.pone.0162176}

\bibitem{LANDAUER1973119}
T.~Landauer, L.~Streeter,
  \href{http://www.sciencedirect.com/science/article/pii/S0022537173800015}{Structural
  differences between common and rare words: Failure of equivalence assumptions
  for theories of word recognition}, Journal of Verbal Learning and Verbal
  Behavior 12~(2) (1973) 119 -- 131.
\newblock \href
  {http://dx.doi.org/https://doi.org/10.1016/S0022-5371(73)80001-5}
  {\path{doi:https://doi.org/10.1016/S0022-5371(73)80001-5}}.
\newline\urlprefix\url{http://www.sciencedirect.com/science/article/pii/S0022537173800015}

\bibitem{Vergara-Martínez2017}
M.~Vergara-Mart{\'i}nez, M.~Comesa{\~{n}}a, M.~Perea,
  \href{https://doi.org/10.3758/s13415-016-0491-7}{The {ERP} signature of the
  contextual diversity effect in visual word recognition}, Cognitive,
  Affective, {\&} Behavioral Neuroscience 17~(3) (2017) 461--474.
\newblock \href {http://dx.doi.org/10.3758/s13415-016-0491-7}
  {\path{doi:10.3758/s13415-016-0491-7}}.
\newline\urlprefix\url{https://doi.org/10.3758/s13415-016-0491-7}

\bibitem{adelman2006}
J.~S. Adelman, G.~D. Brown, J.~F. Quesada, Contextual diversity, not word
  frequency, determines word-naming and lexical decision times, Psychological
  Science 17~(9) (2006) 814--823.
\newblock \href {http://dx.doi.org/10.1111/j.1467-9280.2006.01787.x}
  {\path{doi:10.1111/j.1467-9280.2006.01787.x}}.

\bibitem{Cover2006a}
T.~M. Cover, J.~A. Thomas, Elements of information theory, Wiley, New York,
  2006, 2nd edition.

\bibitem{Brown1966a}
R.~Brown, D.~McNeill, The ``tip of the tongue'' phenomenon, Journal of Verbal
  Learning and Verbal Behavior 5~(4) (1966) 325 -- 337.

\bibitem{Connine1990a}
C.~M. Connine, J.~Mullennix, E.~Shernoff, J.~Yelen, Word familiarity and
  frequency in visual and auditory word recognition, Journal of Experimental
  Psychology: Learning, Memory and Cognition 16 (1990) 1084--1096.

\bibitem{Tuldava1996}
J.~Tuldava, The frequency spectrum of text and vocabulary, J. Quantitative
  Linguistics 3~(1) (1996) 38--50.

\bibitem{Naranan1992c}
S.~Naranan, V.~K. Balasubrahmanyan, Information theoretic models in statistical
  linguistics - {Part II: Word} frequencies and hierarchical structure in
  language., Current Science 63 (1992) 297--306.

\bibitem{Moreno2016a}
I.~Moreno-S{\'a}nchez, F.~Font-Clos, A.~Corral, Large-scale analysis of
  {Zipf's} law in {English} texts, PLOS ONE 11~(1) (2016) 1--19.

\bibitem{Rowland2006a}
C.~F. Rowland, S.~L. Fletcher, The effect of sampling on estimates of lexical
  specificity and error rates, Journal of Child Language 33 (2006) 859--877.

\bibitem{Theakston2001a}
A.~L. Theakston, E.~V.~M. Lieven, J.~M. Pine, C.~F. Rowland, The role of
  performance limitations in the acquisition of verb-argument structure: an
  alternative account, Journal of Child Language 28 (2011) 127--152.

\bibitem{Wells1981a}
C.~G. Wells, Learning through interaction: the study of language development,
  Cambridge University Press, Cambridge, UK, 1981.

\bibitem{Bloom1974a}
L.~Bloom, L.~Hood, P.~Lightbown, Imitation in language development: If, when
  and why, Cognitive Psychology 6 (1974) 380--420.

\bibitem{Bloom1975a}
L.~Bloom, P.~Lightbown, L.~Hood, M.~Bowerman, M.~Maratsos, M.~P. Maratsos,
  Structure and variation in child language, Monographs of the Society for
  Research in Child Development (Serial no. 160) 40~(2) (1975) 1--97.

\bibitem{Bloom1970a}
L.~Bloom, Language development: Form and function in emerging grammars, MIT
  Press, Cambridge, MA, 1970.

\bibitem{Brown1973a}
R.~Brown, A first language: the early stages, Harvard University Press,
  Cambridge, MA, 1973.

\bibitem{Kuczaj1977a}
S.~Kuczaj, The acquisition of regular and irregular past tense forms, Journal
  of Verbal Learning and Verbal Behavior 16 (1977) 589--600.

\bibitem{Childes_Database_Manual_for_American_English}
CHILDES, {American English} Corpora. CHILDES. The Database Manuals. Available
  at http://childes.psy.cmu.edu/manuals/02englishusa.doc. Accessed 17 December
  2012., TalkBank (2012).

\bibitem{Demuth2006a}
K.~Demuth, J.~Culbertson, J.~Alter, Word-minimality, epenthesis, and coda
  licensing in the acquisition of {English}, Language and Speech 49 (2006)
  137--174.

\bibitem{Sachs1983a}
J.~Sachs, Talking about the there and then: the emergence of displaced
  reference in parent-child discourse, in: Children's language, Vol.~4,
  Lawrence Erlbaum Associates, Hillsdale, NJ, 1983, pp. 1--28.

\bibitem{Suppes1974a}
P.~Suppes, The semantics of children's language, American Psychologist 29
  (1974) 103--114.

\bibitem{doi:10.3109/02699209008985472}
G.~Bol, F.~Kuiken,
  \href{http://dx.doi.org/10.3109/02699209008985472}{Grammatical analysis of
  developmental language disorders: {A} study of the morphosyntax of children
  with specific language disorders, with hearing impairment and with {Down}'s
  syndrome}, Clinical Linguistics \& Phonetics 4~(1) (1990) 77--86.
\newblock \href
  {http://arxiv.org/abs/http://dx.doi.org/10.3109/02699209008985472}
  {\path{arXiv:http://dx.doi.org/10.3109/02699209008985472}}, \href
  {http://dx.doi.org/10.3109/02699209008985472}
  {\path{doi:10.3109/02699209008985472}}.
\newline\urlprefix\url{http://dx.doi.org/10.3109/02699209008985472}

\bibitem{Fikkert1994}
P.~Fikkert, \href{http://hdl.handle.net/2066/32125}{On the Acquisition of
  Prosodic Structure}, no.~6, The Hague: Holland Academic Graphics, 1994.
\newline\urlprefix\url{http://hdl.handle.net/2066/32125}

\bibitem{Levelt1994}
C.~Levelt, On the Acquisition of Place, no.~8, The Hague: Holland Academic
  Graphics, 1994.

\bibitem{Bol1995a}
G.~W. Bol, Implicational scaling in child language acquisition: The order of
  production of {Dutch} verb constructions, in: M.~Verrips, F.~Wijnen (Eds.),
  Amsterdam series in child language development: Vol. 3. Papers from the
  {Dutch-German} Colloquium on Language Acquisition, Institute for General
  Linguistics, Amsterdam, 1995, pp. 1--13.

\bibitem{Schaerlaekens1973a}
A.~M. Schaerlaekens, The two-word sentence in child language, Mouton, The
  Hague, 1973.

\bibitem{VanKampen1994a}
J.~{Van Kampen}, The learnability of the left branch condition, in:
  R.~Bok-Bennema, C.~Cremers (Eds.), Linguistics in the Netherlands 1994, John
  Benjamins, Amsterdam/Philadelphia, 1994, pp. 83--94.

\bibitem{Aguirre2000}
C.~Aguirre, La adquisici\'on de las categor\'ias gramaticales en espa\~nol,
  Ediciones de la Universidad Aut\'onoma de Madrid, 2000.

\bibitem{DiezItza1995}
E.~Diez-Itza, Procesos fonol\'ogicos en la adquisici\'on del espa\~nol como
  lengua materna, Actas del XI Congreso Nacional de Ling\"u\'istica Aplicada
  (1995) 225--264.

\bibitem{Hess2003}
K.~Hess~Zimmermann, El desarrollo ling\"u\'istico en los a\~nos escolares:
  an\'alisis de narraciones infantiles, Ph.D. thesis, El Colegio de M\'exico,
  M\'exico (2003).

\bibitem{Jackson1993}
D.~Jackson-Maldonado, D.~Thal, Lenguaje y cognici\'on en los primeros a\~nos de
  vida, Project funded by the John D. and Catherine T. MacArthur Foundation and
  CONACYT (1993).

\bibitem{Thal1993}
D.~Thal, D.~Jackson-Maldonado, Language and cognition in {Spanish}-speaking
  infants and toddlers, Project funded by the John D. and Catherine T.
  MacArthur Foundation (1993).

\bibitem{Linaza1981}
J.~Linaza, M.~E. Sebasti\'an, C.~del Barrio, {Lenguaje, comunicaci\'on y
  comprensi\'on. La adquisici\'on del lenguaje}, Monograf\'ia de Infancia y
  Aprendizaje (1981) 195--198.

\bibitem{Ornat1994}
S.~L. Ornat, A.~Fern\'andez, P.~Gallo, S.~Mariscal, La adquisici\'on de la
  lengua Espa\~nola, Siglo XXI, Madrid, 1994.

\bibitem{Romero1992}
S.~Romero, A.~Santos, D.~Pellicer, The construction of communicative competence
  in {Mexican Spanish} speaking children (6 months to 7 years), Mexico City:
  University of the Americas (1992).

\bibitem{Shiro1997}
M.~Shiro, Getting the story across: A discourse analysis approach to evaluative
  stance in {Venezuelan} children's narratives, unpublished Doctoral
  Dissertation. Harvard University (1997).

\end{thebibliography}

\end{document}